\documentclass[journal]{IEEEtran}

\usepackage{amsmath,amssymb,bm}
\usepackage{array}
\ifCLASSOPTIONcompsoc
  \usepackage[caption=false,font=normalsize,labelfont=sf,textfont=sf]{subfig}
\else
  \usepackage[caption=false,font=footnotesize]{subfig}
\fi
\usepackage{url}
\usepackage{cite}
\usepackage{algorithm}
\usepackage{algorithmic}
\usepackage{color}
\usepackage{graphicx}
\usepackage{multirow}
\usepackage{mathtools}

\usepackage[breaklinks=true,bookmarks=true]{hyperref}

\begin{document}

\title{Single-Perspective Warps in Natural Image Stitching}

\author{Tianli Liao\IEEEauthorrefmark{1} and Nan Li\IEEEauthorrefmark{1}\IEEEauthorrefmark{2}
	\thanks{\IEEEauthorrefmark{1}Equal contribution}
	\thanks{\IEEEauthorrefmark{2}Correspondence author}
	\thanks{T. Liao is with the Center for Combinatorics, Nankai University, Tianjin 300071, China. Email: liaotianli@mail.nankai.edu.cn.}
	\thanks{N. Li is with the Center for Applied Mathematics, Tianjin University, Tianjin 300072, China. E-mail: nan@tju.edu.cn.}
}

\maketitle

\begin{abstract}
Results of image stitching can be perceptually divided into  single-perspective and multiple-perspective. Compared to the multiple-perspective result, the single-perspective result excels in perspective consistency but suffers from projective distortion. In this paper, we propose two single-perspective warps for natural image stitching. The first one is a parametric warp, which is a combination of the as-projective-as-possible warp and the quasi-homography warp via dual-feature. The second one is a mesh-based warp, which is determined by optimizing a total energy function that simultaneously emphasizes different characteristics of the single-perspective warp, including alignment, naturalness, distortion and saliency. A comprehensive evaluation demonstrates that the proposed warp outperforms some state-of-the-art warps, including homography, APAP, AutoStitch, SPHP and GSP.
\end{abstract}

\begin{IEEEkeywords}
	Natural image stitching, image warping, single-perspective, mesh deformation.
\end{IEEEkeywords}

\IEEEpeerreviewmaketitle

\section{Introduction}

\IEEEPARstart{I}{mage} stitching is a process of composing multiple images with narrow but overlapping fields of view to create a larger image with a wider field of view  \cite{szeliski2006image}. The first crucial step is to determine a warping function for each image to
transform it into a common coordinate system. The warps are evaluated in three aspects including alignment, distortion and naturalness.

\begin{figure}
	\centering
	\subfloat[Multiple-perspective (GSP~\cite{chen2016natural}).]{
		\includegraphics[width=0.38\textwidth]{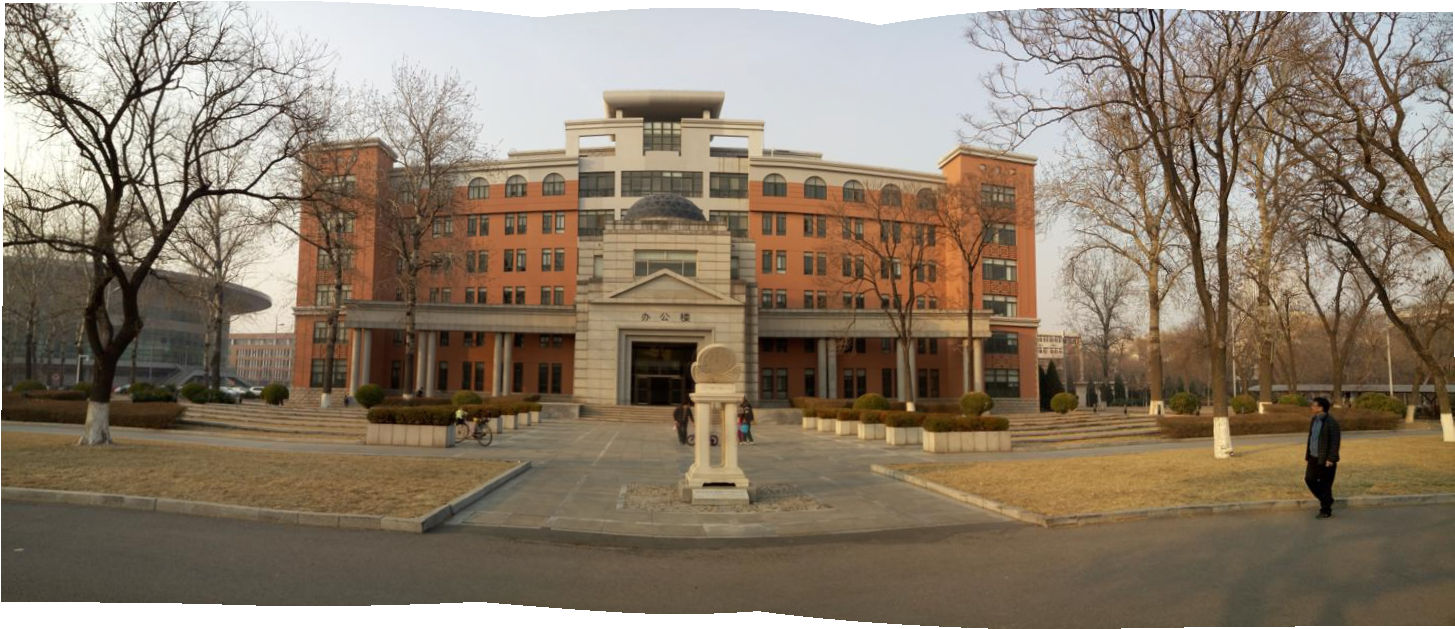}}\\
\vskip-0.5pt
	\subfloat[Single-perspective (APAP~\cite{zaragoza2014projective}).]{
		\includegraphics[width=0.38\textwidth]{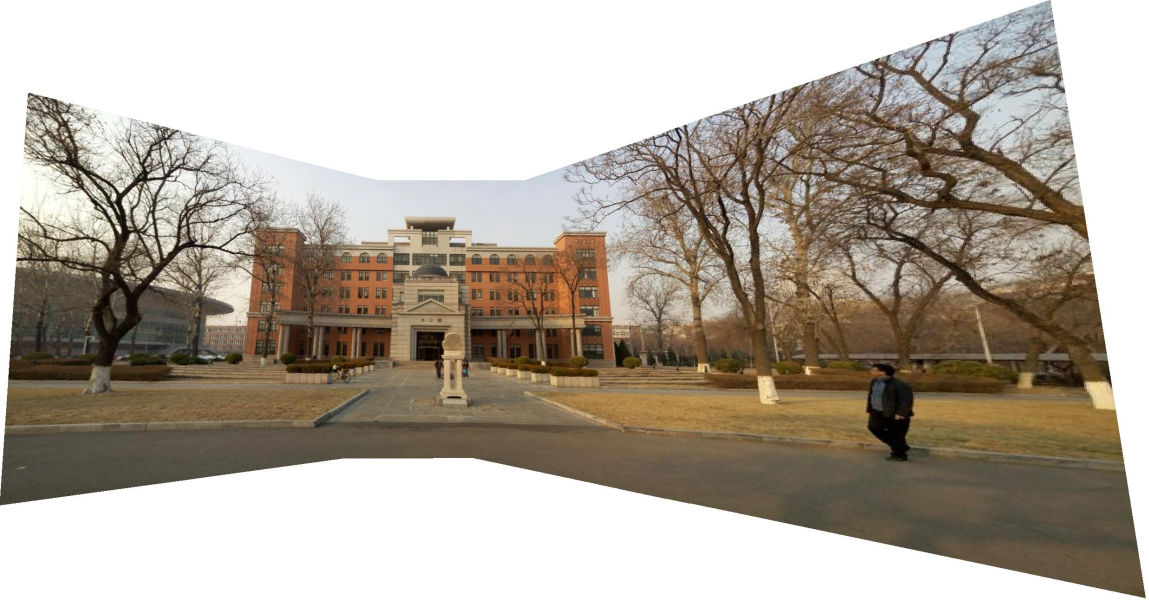}}\\
\vskip-0.5pt
    \subfloat[Single-perspective (Ours).]{
		\includegraphics[width=0.38\textwidth]{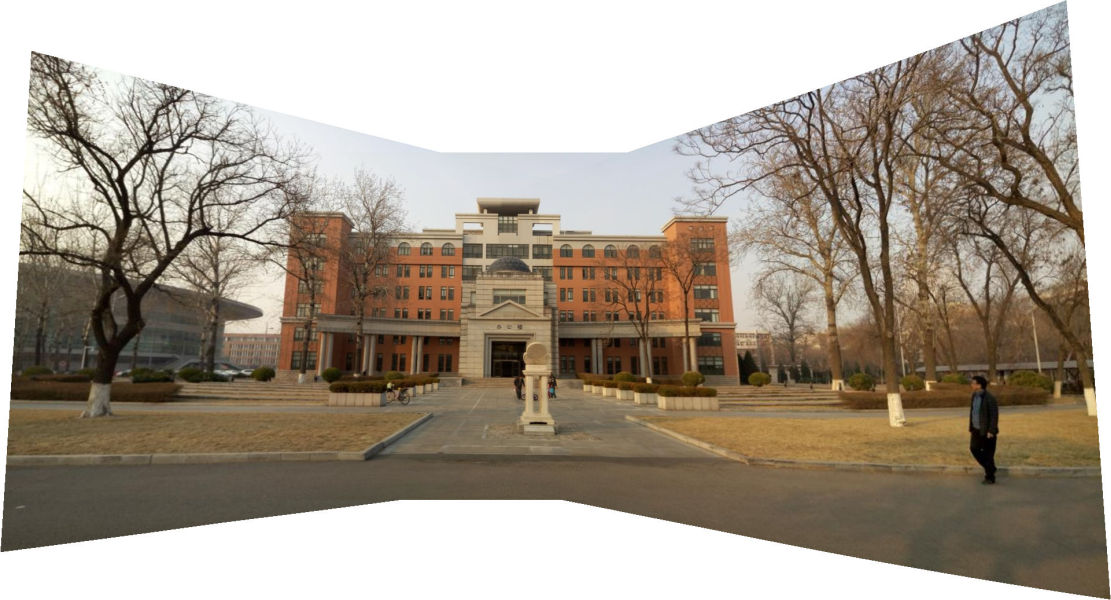}}\\
	\caption{Multiple-perspective warps v.s. single-perspective warps.
}
	\label{fig:1}
\end{figure}

Earlier, the warps focus on addressing the alignment
issue in the overlapping region. The global warps \cite{szeliski1997creating,hartley2003multiple,Brown:2007} are devoted to minimizing the alignment errors between overlapping pixels via one uniform global transformation (mainly homography), which are robust but often not flexible enough. The spatially-varying warps \cite{gao2011constructing,lin2011smoothly,zaragoza2014projective} use multiple local transformations (location dependent) to further improve the alignment accuracy. Instead of minimizing the alignment errors globally, the seam-driven warps \cite{gao2013seam,zhang2014parallax,lin2016seagull} are devoted to finding local overlapping pixels for seamless stitching. Because most of these warps use the homography regularization for smoothly extrapolating the warps into the non-overlapping region, the stitching results are essentially single-perspective, thus they suffer from projective distortion as the homography warp, i.e., shape/area is severely stretched and non-uniformly enlarged (see Figure \ref{fig:1}(b)).

Later, the combination of a warp with better alignment in the overlapping region and a warp with less distortion (mainly similarity) in the non-overlapping region is adopted to address the distortion problem \cite{chang2014shape,lin2015adaptive,li2017quasi}. Because the similarity warp preserves individual perspectives, most of the stitching results are multiple-perspective, therefore they suffer from perspective inconsistency (see Figure \ref{fig:1}(a)). An exception is \cite{li2017quasi} that uses a quasi-homography warp as the warp with less distortion.

Recently, the naturalness issue in the overall stitching result is addressed by guiding the warp either to undergo a global similarity warp that is with minimum line distortion rotation \cite{chen2016natural} or to avoid the extracted
line segments from bending \cite{zhang2016multi}. Because the human eye is more sensitive to lines, emphasizing dual-feature \cite{li2015dual} (point+line) is helpful to
improve the quality of naturalness
for image warping.

In this paper, we propose two single-perspective warps for natural image stitching. The first one is a parametric warp, which is a combination of APAP and QH via DF (see Sec. \ref{para}).
The second one is a mesh-based warp,
which is determined by optimizing a sparse and quadratic total energy function (see Sec. \ref{mesh}).
Implementation details are presented in Sec. \ref{impl} and comprehensive experiments are conducted in Sec. \ref{exp}.

\section{Related Work}
\label{sec2}

\subsection{Warps for Better Alignment}

Conventional stitching methods always employ global warps such like similarity, affine and homography to align images in the overlapping region \cite{hartley2003multiple}, which are robust but often not flexible enough to provide accurate alignment. Gao \textit{et al.} \cite{gao2011constructing} proposed a dual-homography (DH) warp to address the scene with two dominant planes by a weighted sum of two homographies. Lin \textit{et al.} \cite{lin2011smoothly} proposed a smoothly varying affine (SVA) warp, which replaces the global affine warp by a smoothly affine stitching field. Zaragoza \textit{et al.} \cite{zaragoza2014projective} proposed an as-projective-as-possible (APAP) warp in the moving DLT framework, which is able to fine-tune the global homography warp to accommodate location dependent alignment. Other methods combine image alignment with seam-cutting \cite{boykov2001fast,agarwala2004interactive,kwatra2003graphcut} to find a locally aligned region that can be seamlessly blended instead of aligning the entire overlapping region globally. Gao \textit{et al.} \cite{gao2013seam} proposed a seam-driven approach that finds a homography warp with minimal seam cost instead of minimal alignment error.
Zhang and Liu \cite{zhang2014parallax} proposed a parallax-tolerant warp, which combines homography and content-preserving warps to locally align images. Lin \textit{et al.} \cite{lin2016seagull} proposed a seam-guided local alignment approach that iteratively improves the warp by adaptive feature weighting according to the distance to current seams.

\subsection{Warps for Less Distortion}
Many efforts have been devoted to mitigating distortion in the non-overlapping region. A pioneering work \cite{Brown:2007} uses spherical or cylindrical warps to produce multi-perspective results to address this problem, but it necessarily curves straight lines. Chang \textit{et al.} \cite{chang2014shape} proposed a shape-preserving half-projective (SPHP) warp, which
spatially combines a homography warp and a similarity warp such that
it maintains good alignment in the overlapping region while it keeps original perspectives in the non-overlapping region.
Lin and Pankanti \cite{lin2015adaptive} proposed an adaptive as-natural-as-possible (AANAP) warp,
which combines a linearized homography warp and a global similarity warp with the smallest rotation angle such that the stitching result is more natural-looking.
Li \textit{et al.} \cite{li2017quasi} proposed a quasi-homography (QH) warp to balance perspective distortion against
projective distortion in the non-overlapping region, which creates a single-perspective stitching result.

\subsection{Warps for Better Naturalness}
Recently, some stitching methods model image warping with mesh deformation, which are obtained via energy minimization. The naturalness issue is addressed by emphasizing line features.
Chen \textit{et al.} \cite{chen2016natural}
proposed a global-similarity-prior (GSP) warp, which constrains the warp to undergo a global
similarity warp with minimum line distortion rotation.
Zhang \textit{et al.} \cite{zhang2016multi} proposed a warp
to produce an orthogonal projection of a wide-baseline scene, which constrains the warp to preserve extracted line segments.
Li \textit{et al.} \cite{li2015dual} proposed a warp based on dual-feature (DF), which not only
improves the alignment accuracy in low-texture cases but
also prevents some undesired distortion.

\section{Single-Perspective Parametric Warp}\label{para}

The homography warp is the most classic single-perspective warp for image stitching, but it carries the motion assumption that the images are taken by purely rotational camera motion or the scene assumption that the geometry of the scene is planar or effectively planar, such as the scene is sufficiently far away.
If such conditions are not satisfied, in other words, the camera motion involves translation and the scene involves non-planar geometry, the homography warp inevitably introduces alignment, distortion and naturalness issues in the panorama. In fact, under the general assumption of motion and scene, the homography warp relates the images of the same scene from different views by a common dominant plane in the sense of ``least squares''.
In this section, we first describe an analysis of single-perspective warps in aspects of alignment, distortion and naturalness with regard to the common dominant plane, then we propose a parametric warp to address the three issues.

\subsection{Mathematical Setup}

Let $I$ and $I^{\prime}$ denote the \emph{target} image and the \emph{reference} image respectively.
A homography warp $\mathcal{H}$ is a planar transformation, which relates pixel coordinates $(x,y)\in I$ to $(x',y')\in I^{\prime}$ by
\begin{equation}
 \left\{\begin{array}{c}
         x'=f(x,y) \\
         y'=g(x,y)
       \end{array}
 \right.,
\end{equation}
where
\begin{equation}\label{eq_fg}
f(x,y)=\frac{h_1x+h_2y+h_3}{h_7x+h_8y+1},
g(x,y)=\frac{h_4x+h_5y+h_6}{h_7x+h_8y+1}.
\end{equation}

\subsection{Alignment Issue}

\begin{figure*}
	\centering
	\includegraphics[width=0.8\textwidth]{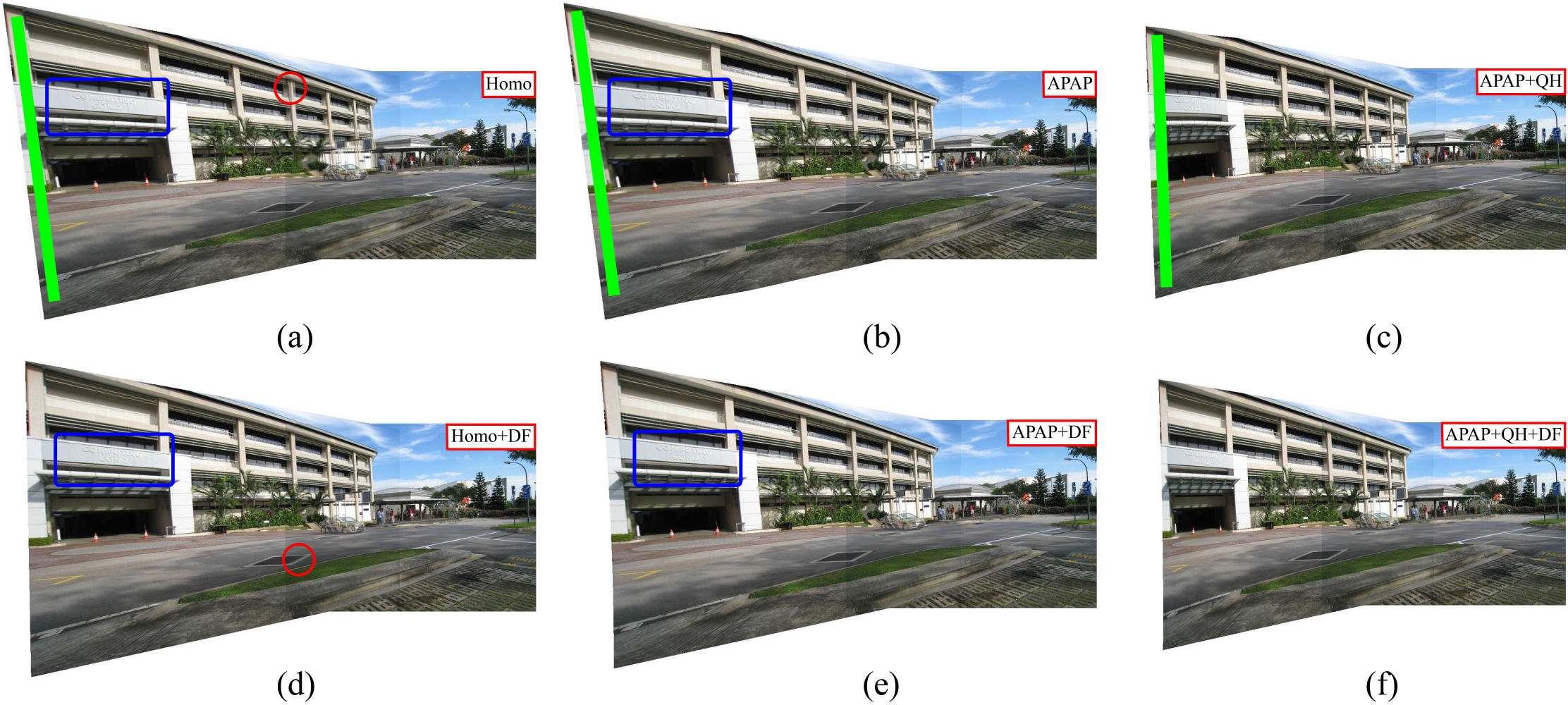}\\
	\caption{Comparison of single-perspective parametric warps. (a) Homography (b) APAP (c) APAP$+$QH (d) Homography+DF (e) APAP+DF (f) APAP+QH+DF. Alignment, naturalness, distortion issues are highlighted in {\color{red}red}, {\color{green}green} and {\color{blue}blue}, which are addressed by APAP, DF and QH respectively. (Best to zoom-in and view on screen)}
	\label{fig:six}
\end{figure*}

Conventionally, $\mathcal{H}$ is estimated from a set of feature correspondences via direct linear transformation (DLT). Suppose given feature correspondences are from a planar geometry of the scene, then $\mathcal{H}$ is the accurate homography between them from different views. If they are from a non-planar geometry, then $\mathcal{H}$ is the approximate homography that relates $I$ to $I^{\prime}$ by a common dominant plane, which fits given feature correspondences in the sense of ``least squares''.
Intuitively, the alignment quality is more accurate for the content that is more near the plane and is poorer for the content that is more off the plane (see Figure \ref{fig:six}(a)).

In contrast to using a single global homography warp to align the images, APAP \cite{zaragoza2014projective} uses a spatially-varying warp $\mathcal{H}_{*}$ that consists of multiple local homography warps for image alignment. In other words, instead of using the common dominant plane to approximate the non-planar geometry of the scene, APAP uses the field of spatially-varying planes in the overlapping region, which evolves to the
common dominant plane in the non-overlapping region. The approach is called moving DLT, which has been proven very effective for image alignment, especially in cases of general motion and non-planar scene. It is worth to note that APAP still creates a single-perspective stitching result (see Figure \ref{fig:six}(b)).

\subsection{Naturalness Issue}\label{sec_df}

Suppose the geometry of the scene is effectively planar, then $\mathcal{H}$ creates a natural-looking panorama. For the fundamental two-image stitching problem, it means that $\mathcal{H}$ transforms $I$'s perspective into $I'$' such that $I$ is warped as an extended view of $I'$ without violating any relative geometry. If the geometry of the scene is non-planar, then the panorama created by $\mathcal{H}$ may be less natural-looking, which can be evidently observed from the line orientation (see Figure \ref{fig:six}(a)). $\mathcal{H}_{*}$ may also suffer the naturalness issue, since an invisible misalignment of lines in the overlapping region can be amplified into a visible violation of relative geometry in the overall stitching result (see Figure \ref{fig:six}(b)).

In fact,
considering dual-feature \cite{li2015dual} in feature correspondences is helpful to
improve the naturalness quality.
In contrast to estimating the warp via only point features, $\mathcal{H}$ \cite{li2015dual} (or $\mathcal{H}_{*}$ \cite{joo2015line}) via DF
not only emphasizes the alignment quality in the overlapping region but also stresses the naturalness quality in the the overall stitching result (see Figure \ref{fig:six}(d,e)).
In other words, the common dominant plane more protects lines from being misaligned
in the overlapping region.

\subsection{Distortion Issue}\label{sec_quasi}

Besides the alignment and naturalness issues, $\mathcal{H}$ and $\mathcal{H}_{*}$ may still suffer the distortion issue in the non-overlapping region. Intuitively, the distortion is negligible for the content that is near the common dominant plane, but it gets worse for the content that is off the plane (see Figure \ref{fig:six}(a,b,d,e)).

\begin{figure}
	\centering
	\includegraphics[width=0.45\textwidth]{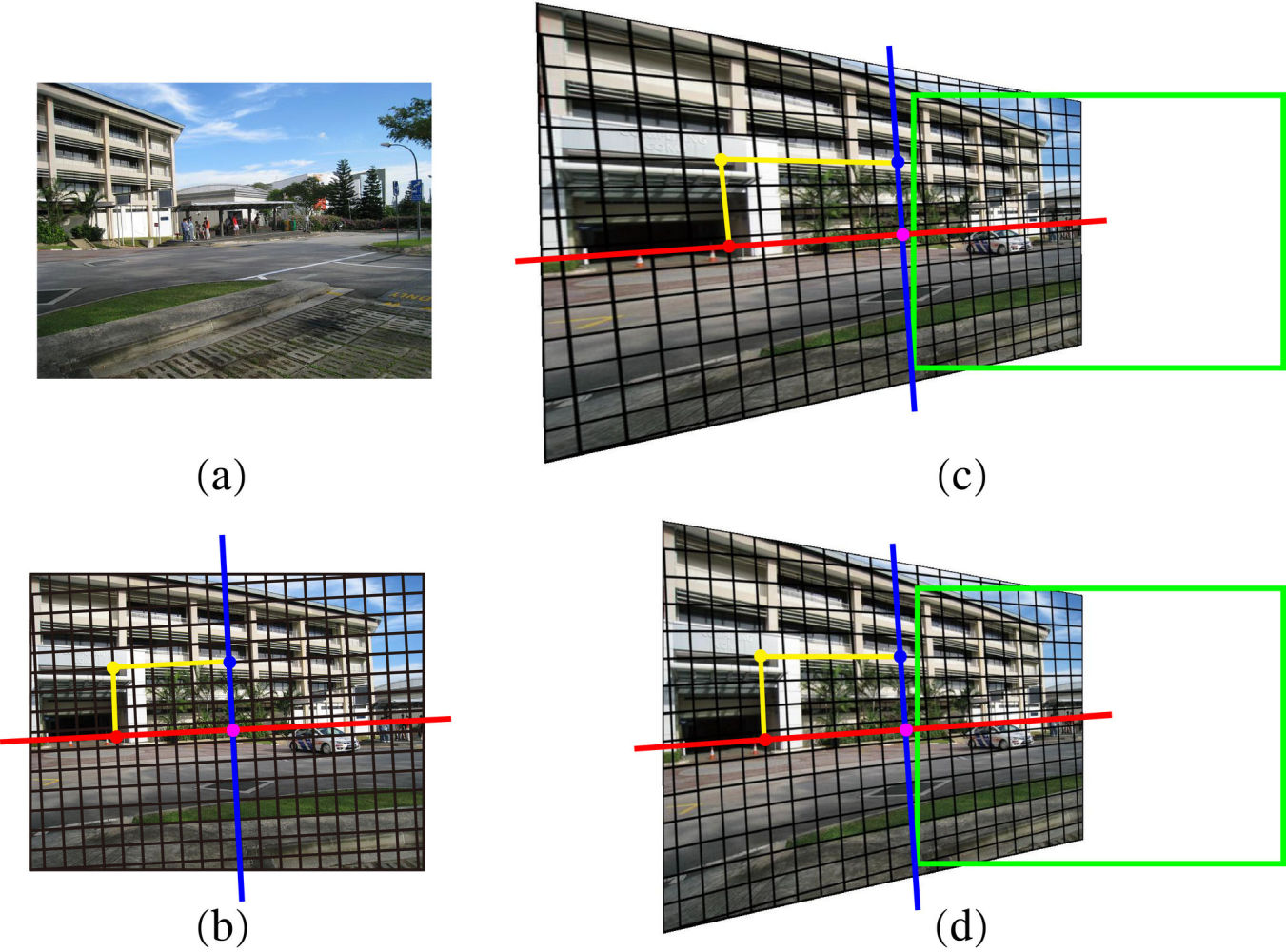}\\
	\caption{Sketch of quasi-homography warps. (a) Reference. (b) Meshed target. (c) Meshed result of homography. (d) Meshed result of quasi-homography.}
	\label{fig:mesh}
\end{figure}

In contrast to using the homography regularization as $\mathcal{H}$ and $\mathcal{H}_{*}$, QH \cite{li2017quasi} uses a quasi-homography regularization $\mathcal{H}_{\dag}$, which can mitigate the distortion and preserve the perspective. Recall that $\mathcal{H}_{\dag}$ can be formulated as the solution of a bivariate system
\begin{align}
\frac{y^{\prime}-g(\Pi_1(x,y))}{x^{\prime}-f(\Pi_1(x,y))}&=s(x,y,k_1),\label{newQH1}\\
\frac{y^{\prime}-g_{\dag}(\Pi_2(x,y))}{x^{\prime}-f_{\dag}(\Pi_2(x,y))}&=s(x,y,k_2).\label{newQH2}
\end{align}
Figure \ref{fig:mesh}(b) shows the sketch of QH, where the points $\Pi_1(x,y)$ and $\Pi_2(x,y)$ (in {\color{red}red} and {\color{blue}blue}) are the projections of $(x,y)$ (in \textcolor{yellow}{yellow}) onto the cross-lines $l_u$ and $l_v$ (in {\color{red}red} and {\color{blue}blue}) which intersect at $(x_{\ast},y_{\ast})$ (in \textcolor[rgb]{1,0,1}{magenta}). In addition, $k_1\cdot k_2=-1$, $f_{\dag}(x,y),g_{\dag}(x,y)$ are the first order truncations of the Taylor's series of $f,g$ (\ref{eq_fg}) at $(x_{*},y_{*})$, $s(x,y,k)$ is the slope of the line in $I'$ corresponding to the line passing $(x,y)$ with slope $k$ in $I$, which is calculated by
\begin{equation}
s(x,y,k)=\frac{g_x(x,y)+kg_y(x,y)}{f_x(x,y)+kf_y(x,y)},
\end{equation}
where $f_x,f_y,g_x,g_y$ denote the partial derivatives of $f,g$.

In \cite{li2017quasi}, by assuming that the images are taken by oriented cameras via horizontal motion, $l_u$ is set to be the horizontal line
that remains horizontal after warping and $l_v$ is set to be the vertical partition line that is closest to the border
of the overlapping region and the non-overlapping region. In fact, if a homography warp is given, there exists a unique family of parallels that remains parallels after warping. Their slope can be determined by solving $s_x=s_y=0$,
where $s_x,s_y$ denote the partial derivatives of $s(x,y,k)$.
The slopes of the parallels before and after warping are
\begin{equation}\label{lv}
k_1=-\frac{h_7}{h_8},~~~s(x,y,k_1)= \frac{h_4 h_8-h_5 h_7}{h_1 h_8-h_2 h_7}.
\end{equation}
Therefore, $l_v$ can be set to be the partition line with slope $k_1$ that is closest to the border
of the overlapping region and the non-overlapping region, and $l_u$ can be set to be the line that is orthogonal with $l_v$ before and after warping, i.e.,
\begin{equation}\label{lu}
  k_1\cdot k_2=-1,~~~s(x,y,k_2)\cdot s(x,y,k_1)=-1.
\end{equation}

In fact, (\ref{lv},\ref{lu}) help QH get rid of the assumption of camera orientation and motion. Furthermore, the ratios of lengths on the family of parallels that parallel to $l_v$ are preserved. In other words, the directional derivative of $\mathcal{H}_{\dag}$ along each line with slope $k_1$ is a constant, which coincides with the change of coordinates technique in \cite{chang2014shape}. We use the analysis to characterize the distortion issue for the single-perspective mesh deformation (see Sec. \ref{sec_dist}).

\subsection{Composite Warp}
Composing the QH warp $\mathcal{H}_{\dag}$ with the APAP warp $\mathcal{H}_{*}$ by
\begin{equation}
  \mathcal{H}_{\dag}\circ\mathcal{H}^{-1}\circ\mathcal{H}_{*},
\end{equation}
creates a more natural-looking single-perspective result (see Figure \ref{fig:six}(c,f)), where $\mathcal{H}$ is the homography warp that is extrapolated from $\mathcal{H}_{*}$ into $\mathcal{H}_{\dag}$.

\section{Single-Perspective Mesh Deformation}\label{mesh}

The combination of APAP, DF and QH normally creates a natural-looking single-perspective stitching result, but the warp could still bend some salient lines (see Figure \ref{fig-visible}(a)).
In order to address such issue, we propose a mesh-based warp, which regards the alignment, naturalness, distortion, saliency issues as different terms of energies, then minimizes the total energy function to get the desired single-perspective warp.

\subsection{Mathematical Setup}

Let $I$ and $I^{\prime}$ denote the \emph{target} image and the \emph{reference} image respectively.
After building mesh grids for $I$ and indexing grid vertices from $1$ up to $n$, we reshape the $n$ vertices into a $2n$-dimension vector
$V=[x_1~~y_1~~x_2~~y_2~~\ldots~~x_n~~y_n]^{\rm{T}}$,
then the corresponding $n$ vertices after mesh deformation are formed into $\hat{V}=[\hat{x}_1~~\hat{y}_1~~\hat{x}_2~~\hat{y}_2~~\ldots~~\hat{x}_n~~\hat{y}_n]^{\rm{T}}$.

\begin{figure}
	\centering
	\includegraphics[width=0.3\textwidth]{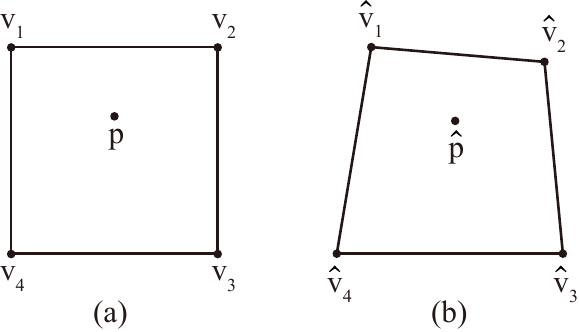}
	\caption{Bilinear interpolation of sample points. (a)(b) Grid vertices and sample point before and after deformation. }\label{fig_align}
\end{figure}

Similar to~\cite{liu2013bundled}, for any sample point $p$ in $I$, we characterize it as a bilinear interpolation of its four enclosing grid vertices $v_1,v_2,v_3,v_4$, i.e.,
\begin{equation}
\varphi(p)=w_1v_1+w_2v_2+w_3v_3+w_4v_4.
\end{equation}
By assuming the coefficients are fixed, then the corresponding point $\hat{p}$ is characterized as the bilinear interpolation $\varphi(\hat{p})=w_1\hat{v}_1+w_2\hat{v}_2+w_3\hat{v}_3+w_4\hat{v}_4$ (see Figure \ref{fig_align}). Consequently, any constraint on the point correspondences can be expressed as a constraint on the vertex correspondences.

After above preparation, we define the total energy function \begin{equation}\label{eq_energy}
E(\hat{V})=E_{\mathrm{p}}(\hat{V})+\lambda_{\mathrm{l}}E_{\mathrm{l}}(\hat{V})+E_{\mathrm{cl}}(\hat{V})+\lambda_{\mathrm{s}}E_{\mathrm{s}}(\hat{V}),
\end{equation}
where $E_{\mathrm{p}}(\hat{V})$ addresses the alignment issue by enhancing the point correspondences, $E_{\mathrm{l}}(\hat{V})$ addresses the naturalness issue by strengthening the line correspondences, $E_{\mathrm{cl}}(\hat{V})$ addresses the distortion issue by rearranging the cross-line correspondences with a homography warp prior and $E_{\mathrm{s}}(\hat{V})$ addresses the saliency issue by protecting salient lines from being bent. A mesh-based warp is determined by solving $\min E(\hat{V})$.

\subsection{Alignment Term}\label{sec_align}

Given the set of point correspondences $\{(p_i,p'_i)\}_{i=1}^{N}$, where $p_i=(x_i,y_i)\in I$ and $p'_i=(x'_i,y'_i)\in I'$ (see Figure \ref{fig:uv_lines}(a)), then
\begin{equation}\label{eq_al}
  E_{\mathrm{p}}(\hat{V})=\sum_{i=1}^{N}\|\varphi(\hat{p}_i)-p'_i\|^2=\|W_{\mathrm{p}}\hat{V}-P\|^2,
\end{equation}
where $P\in\mathbb{R}^{2N}$ consists of coordinates of $p'_i$, $W_{\mathrm{p}}\in\mathbb{R}^{2N\times 2n}$ consists of coefficient vectors of coordinates of the bilinear interpolation of $\hat{p}_i$.

\subsection{Naturalness Term}\label{sec_natr}

\begin{figure*}
	\centering
	\subfloat[Point and line features.]{
		\includegraphics[width=0.24\textwidth]{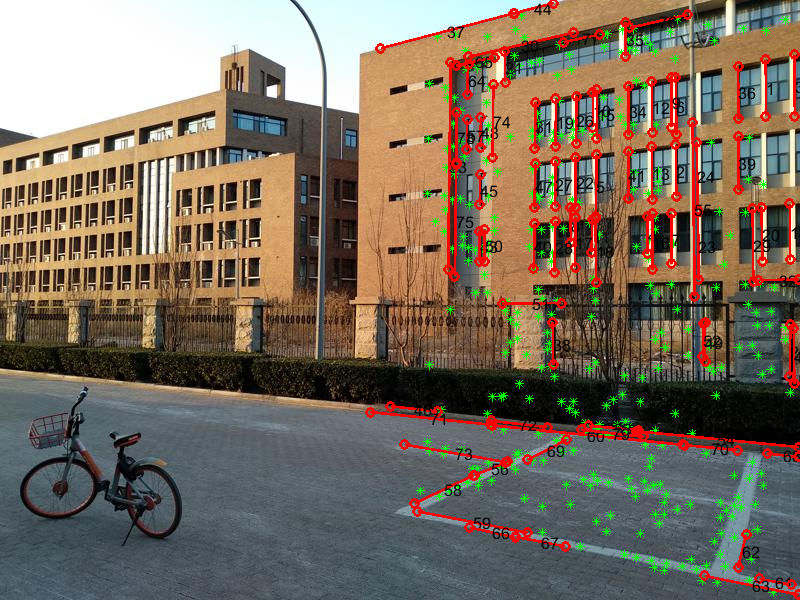}
		\includegraphics[width=0.24\textwidth]{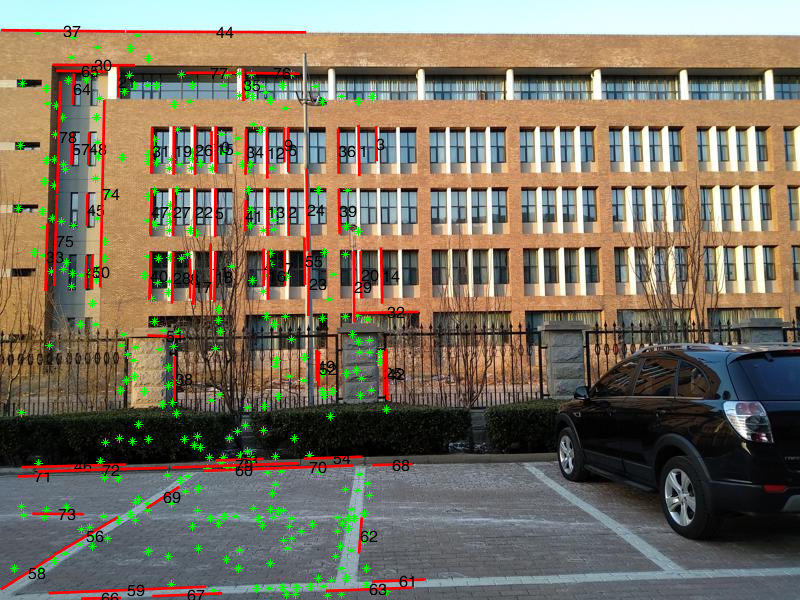}}
	\subfloat[Cross-line features.]{
		\includegraphics[width=0.24\textwidth]{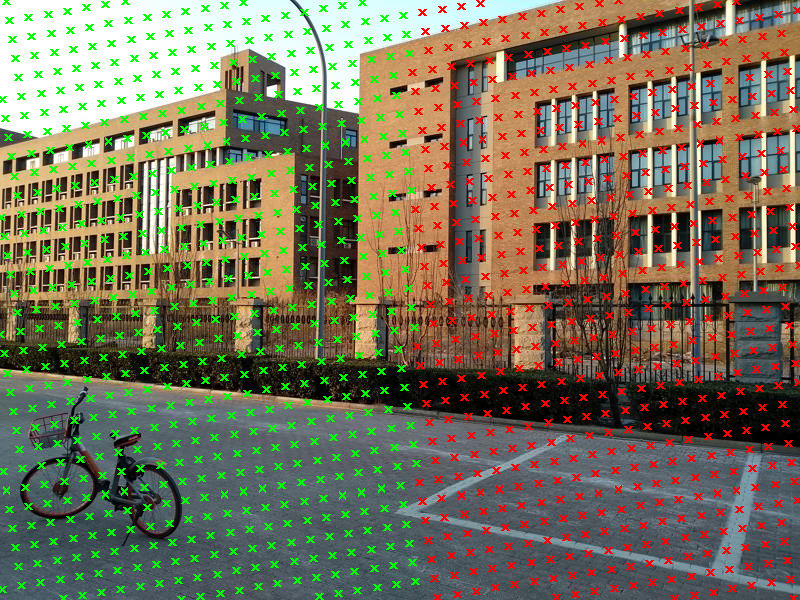}}
	\subfloat[Detected line segments.]{
		\includegraphics[width=0.24\textwidth]{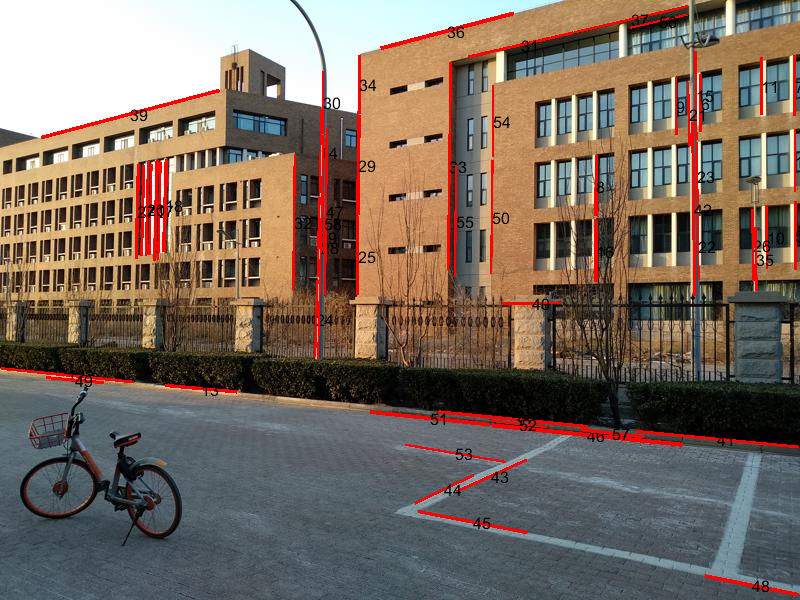}}\\
	\caption{Illustration of sample points in mesh deformation. (a) Point ({\color{green}green}) and line ({\color{red}red}) correspondences in the target (left) and reference (right) image. (b) Uniformly sampled cross-line features (overlapping region in {\color{red}red} and  non-overlapping region in {\color{green}green}) in the target. (c) Detected line segments in the target.}
	\label{fig:uv_lines}
\end{figure*}

Given the set of line correspondences $\{(l_j,l'_j)\}_{j=1}^{M}$, where
$l_j\in I$ is represented by the line segment with the starting point $p_j^s$ and the ending point $p_j^e$, and $l'_j\in I'$ is represented by the line equation $a_jx+b_jy+c_j=0$ (see Figure \ref{fig:uv_lines}(a)), then
\begin{equation}\label{eq_na}
  E_{\mathrm{l}}(\hat{V}) =\sum_{j=1}^{M}|\langle\varphi(\hat{p}_j^{s,e}),\vec{\mathbf{n}}_j\rangle+c_j|^2  =\|W_{\mathrm{l}}\hat{V}+C\|^2,
\end{equation}
where $\vec{\mathbf{n}}_j=(a_j,b_j)^T$, $C=(c_1,c_1,\ldots,c_M,c_M)^T$ and $W_{\mathrm{l}}\in\mathbb{R}^{2M\times 2n}$ consists of coefficient vectors of the inner products of the bilinear interpolation of $\hat{p}_j^{s,e}$ and $\vec{\mathbf{n}}_j$.

\subsection{Distortion Term}\label{sec_dist}

Inspired by the above quasi-homography warp that mitigates the distortion by linearizing the scale on $l_u$ while preserves the perspective by keeping the corresponding slopes of cross-lines that are parallel to $l_u$ and $l_v$, we divide $E_{\mathrm{cl}}(\hat{V})$ into $E_{\mathrm{ps}}(\hat{V})$ that preserves the desired perspective given by a homography warp prior and $E_{\mathrm{pj}}(\hat{V})$ that mitigates the projective distortion.

Given the set of cross-line correspondences $\{(l_i^u,{l'}_i^u)\}_{i=1}^{S}$ and $\{(l_j^v,{l'}_j^v)\}_{j=1}^{T}$, where $l_i^u$ and $l_j^v$ are parallel to $l_u$ and $l_v$ that are calculated from a homography warp prior, then we uniformly sample them with $L_i$ and $K_j$ points $\{p^{u,i}_k\}_{k=1}^{L_i}$ and $\{p^{v,j}_k\}_{k=1}^{K_j}$ (see Figure \ref{fig:uv_lines}(b)), then
\begin{align}\label{eq_ps}
E_{\mathrm{ps}}(\hat{V}) &= \sum_{i=1}^S \sum_{k=1}^{L_i-1}|\langle \varphi(\hat{p}^{u,i}_{k+1})-\varphi(\hat{p}^{u,i}_k),\vec{\mathbf{n}}^u_i\rangle|^2\nonumber\\
              & +\sum_{j=1}^T\sum_{k=1}^{K_j-1}|\langle \varphi(\hat{p}^{v,j}_{k+1})-\varphi(\hat{p}^{v,j}_k),\vec{\mathbf{n}}^v_j\rangle|^2\nonumber\\
              & +\sum_{j=1}^T\sum_{k=1}^{K_j-2}\|\varphi(\hat{p}^{v,j}_k)+\varphi(\hat{p}^{v,j}_{k+2})-2\varphi(\hat{p}^{v,j}_{k+1})\|^2\nonumber\\
              &= \|W_{\mathrm{ps}}\hat{V}\|^2 ,
\end{align}
where ${\vec{\mathbf{n}}}^u_i$ and ${\vec{\mathbf{n}}}^v_j$ are the normal vectors of ${l'}_i^u$ and ${l'}_j^v$, $W_{\mathrm{ps}}\in \mathbb{R}^{(\sum_{i=1}^S(L_i-1)+\sum_{j=1}^T(3K_j-5))\times 2n}$. In fact, the first two terms preserve the slopes of ${l'}_i^u$ and ${l'}_j^v$, and the last term preserves the ratios of
lengths on ${l}_j^v$.

Let $\Omega$ denote the preimage of the non-overlapping region of $I$, then
\begin{align}\label{eq_pj}
E_{\mathrm{pj}}(\hat{V}) & =\sum_{i=1}^S\sum_{k=1}^{|p^{u,i}_k\in\Omega|-2}\|\varphi(\hat{p}^{u,i}_k)+\varphi(\hat{p}^{u,i}_{k+2})-2\varphi(\hat{p}^{u,i}_{k+1})\|^2\nonumber\\
& = \|W_{\mathrm{pj}}\hat{V}\|^2.
\end{align}
where $W_{\mathrm{pj}}\in \mathbb{R}^{\sum_{i=1}^S 2(|p^{u,i}_k\in\Omega|-2)\times 2n}$.
In fact, (\ref{eq_pj}) linearizes the scale on ${l}_i^u$ in $\Omega$.

In summary,
\begin{equation}\label{eq_di}
  E_{\mathrm{cl}}(\hat{V})=\lambda_{\mathrm{ps}}E_{\mathrm{ps}}(\hat{V})+\lambda_{\mathrm{pj}}E_{\mathrm{pj}}(\hat{V}).
\end{equation}

\subsection{Saliency Term}\label{sec_sal}

Given the set of salient lines $\{l^s_k\}_{k=1}^{Q}$, where each $l^s_k\in I$ is uniformly sampled with $J_k$ points $\{p^k_j\}_{j=1}^{J_k}$ (see Figure \ref{fig:uv_lines}(c)), then
\begin{equation}\label{eq_sal}
E_{\mathrm{s}}(V) =\sum_{k=1}^{Q}\sum_{j=1}^{J_k-1}\|\langle \varphi(\hat{p}^k_{j+1})-\varphi(\hat{p}^k_j),\vec{\mathbf{n}}_k \rangle\|^2
=\|W_{\mathrm{s}}\hat{V}\|^2,
\end{equation}
where $\vec{\mathbf{n}}_k$ is the normal vector of ${l'}^s_k\in I'$ that is calculated from the homography warp prior and $W_{\mathrm{s}}\in \mathbb{R}^{\sum_{k=1}^{Q}(J_k-1)\times 2n}$.

\begin{figure*}
	\centering
    \subfloat[Result of APAP$+$DF$+$QH.]{
		\includegraphics[width=0.22\textwidth]{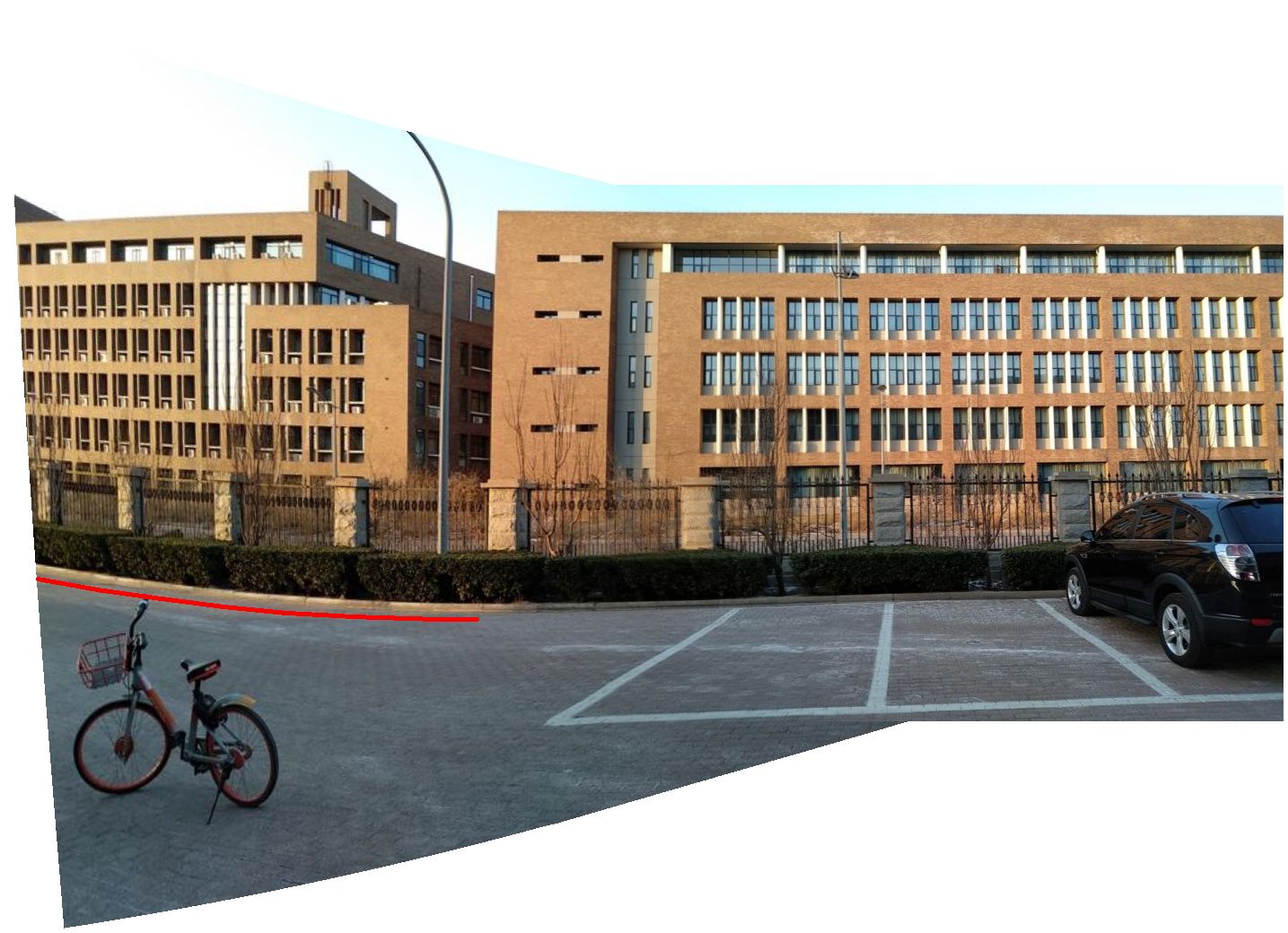}}
	\hskip 4pt
	\subfloat[Result of mesh deformation with $\lambda_s=0$.]{
		\includegraphics[width=0.22\textwidth]{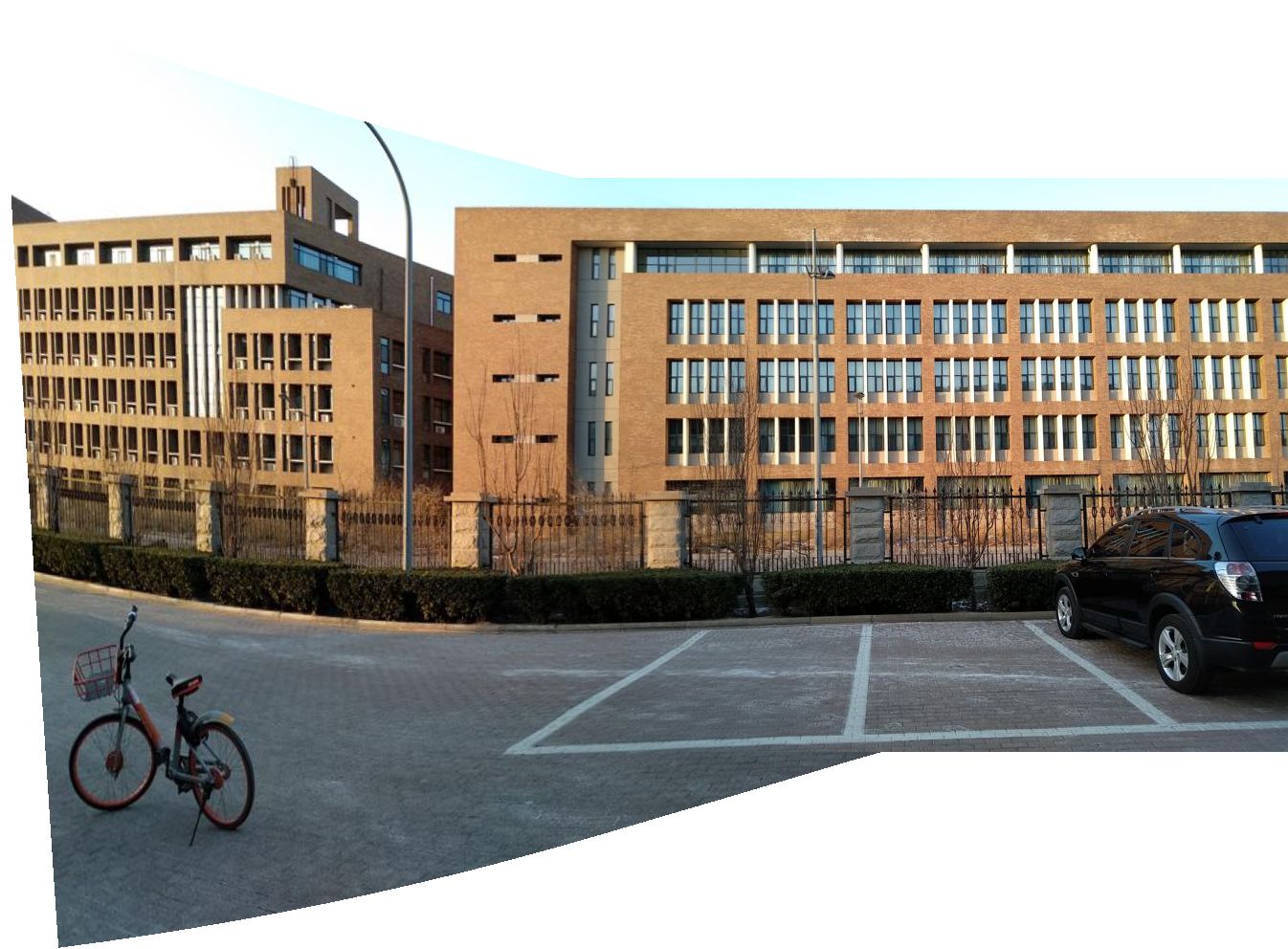}}
	\hskip 4pt
    \subfloat[Result of mesh deformation with $\lambda_s=5$.]{
		\includegraphics[width=0.22\textwidth]{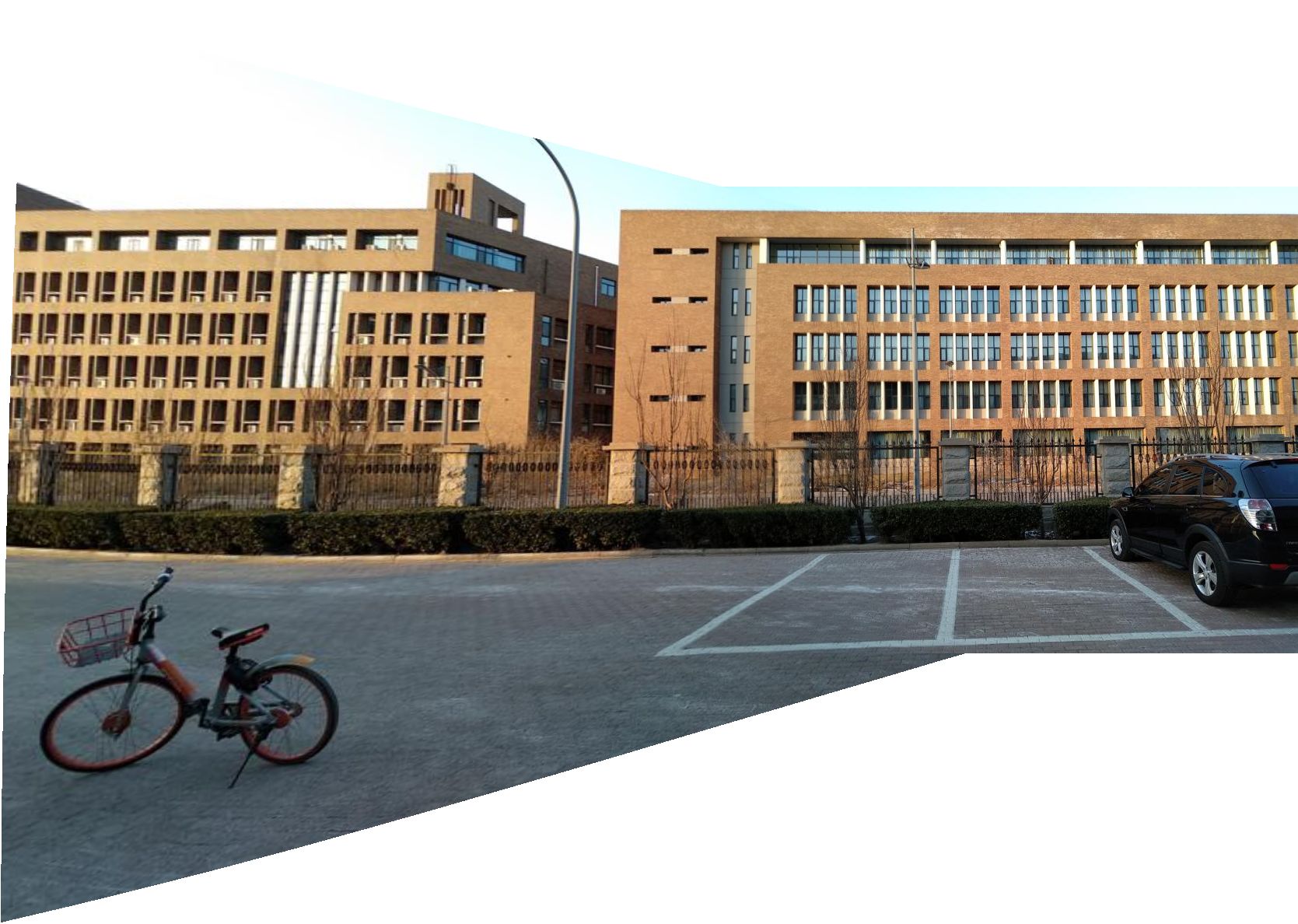}}
	\hskip 4pt
	\subfloat[Result of mesh deformation with $\lambda_s=5000$.]{
		\includegraphics[width=0.22\textwidth]{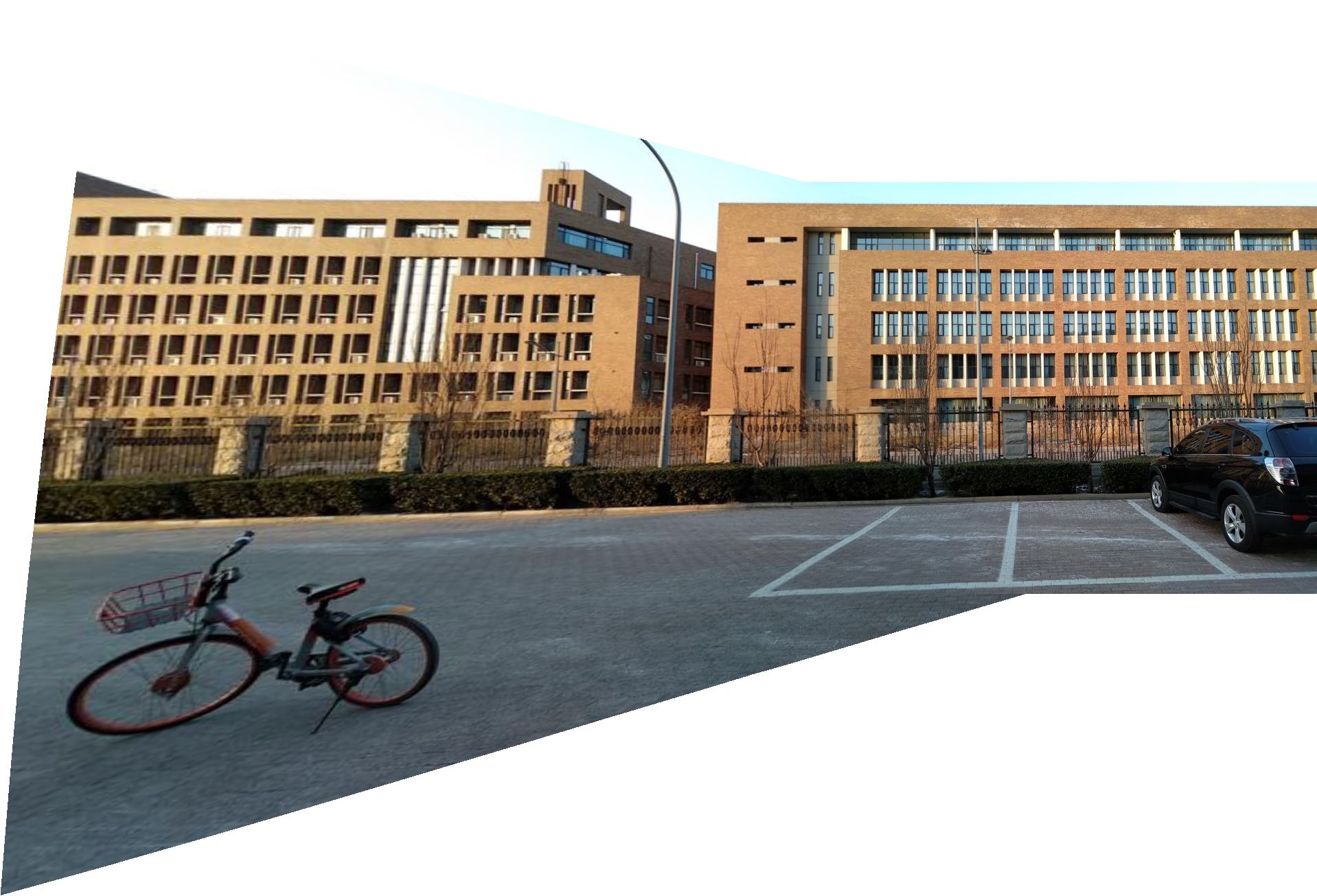}}\\
	\caption{Comparison of stitching results on protecting salient lines. If $\lambda_{\mathrm{s}}$ is too small, then salient lines are obviously bent (close to the result of APAP$+$DF$+$QH). If $\lambda_{\mathrm{s}}$ is too large, then the projective distortion becomes serious (close to the result of homography). An appropriate trade-off of $\lambda_{\mathrm{s}}$ will reach the balance.}\label{fig-visible}
\end{figure*}

Figure \ref{fig-visible} shows a comparison example with different values of $\lambda_s$, which demonstrates the capability of the saliency term for protecting salient lines.

\subsection{Optimization}

Because all terms are quadratic, we reform the total energy function (\ref{eq_energy}) by
\begin{align}\label{eq_newenergy}
E(\hat{V})=\left\|
\begin{bmatrix}
W_{\mathrm{p}} \\ \sqrt{\lambda_{\mathrm{l}}}W_{\mathrm{l}}\\ \sqrt{\lambda_{\mathrm{ps}}}W_{\mathrm{ps}}\\ \sqrt{\lambda_{\mathrm{pj}}}W_{\mathrm{pj}}\\ \sqrt{\lambda_{\mathrm{s}}}W_{\mathrm{s}}
\end{bmatrix}
\hat{V} -
\begin{bmatrix}
P\\ -\sqrt{\lambda_{\mathrm{l}}}C \\ 0\\ 0\\ 0
\end{bmatrix}
\right\|^2.
\end{align}
$\min E(\hat{V})$ can be efficiently sloved by any sparse linear solver.

\section{Implementation}\label{impl}

\subsection{Two-Image Stitching}

Given a pair of two images, we first estimate a homography warp using both point and line correspondences that transforms the perspective of the target into that of the reference, then the calculated cross-line correspondences and the detected line segments are uniformly sampled in the target image, finally we optimize the total energy function to obtain the warped results. A brief algorithm is given in Algorithm \ref{algor_1}.

\begin{algorithm}
	\caption{Two-image stitching.}
	\label{algor_1}
	\begin{algorithmic}[1]
		\REQUIRE a target image $I$ and a reference image $I'$.
		\ENSURE a stitched image.
		\STATE Match point and line features between $I$ and $I'$ to obtain $\{(p_i, p'_i)\}$ and $\{(l_j, l'_j)\}$.
		\STATE Calculate a homography warp $\mathcal{H}$ via dual-feature.
		\STATE Calculate $\{(l_i^u,{l'}_i^u)\}$ and $\{(l_j^v,{l'}_j^v)\}$ from $\mathcal{H}$ by (\ref{lv},\ref{lu}).
        \STATE Extract salient line segments in $I$ to obtain $\{l_k^s\}$.
		\STATE Uniformly sample $\{l_i^u\}$, $\{l_j^v\}$ and $\{l_k^s\}$. 
		\STATE Solve $\hat{V}$ via minimizing the total energy function (\ref{eq_newenergy}). 
		\STATE Warp $I$ via the bilinear interpolation with respect to $\hat{V}$ and composite the warped result with $I'$ via linear blending.
	\end{algorithmic}
\end{algorithm}

\subsection{Multiple-Image Stitching}

\begin{figure}
	\centering
	\subfloat[Result w/o line correspondences.]{
	\includegraphics[width=0.45\textwidth]{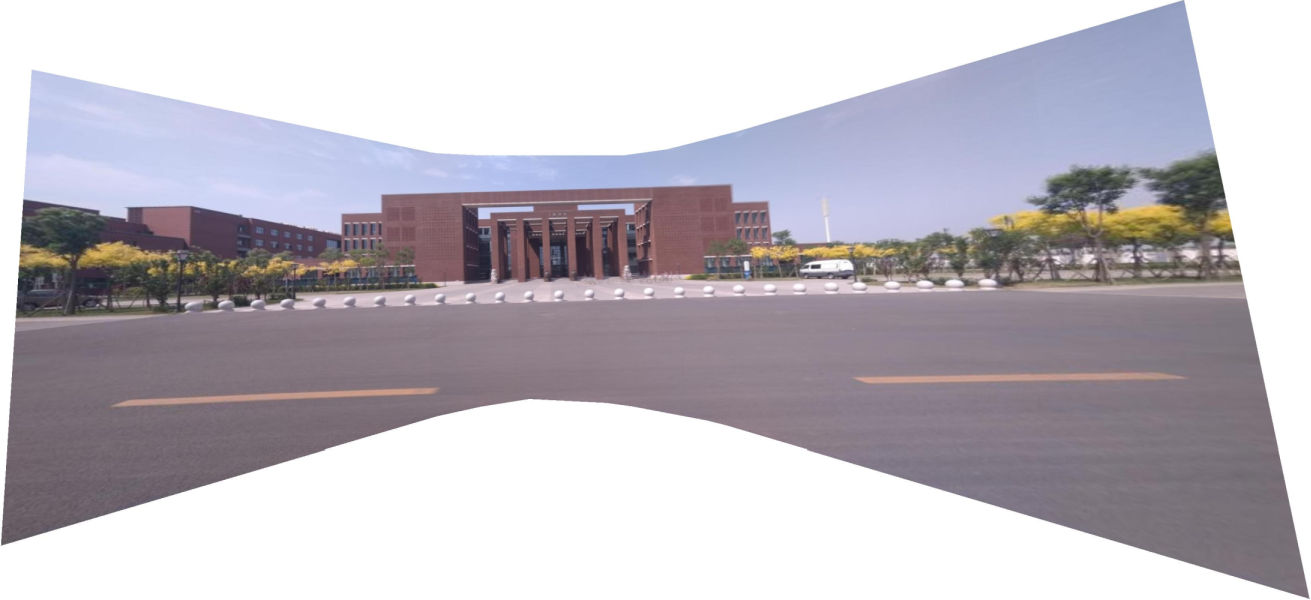}}\\
	\vspace{-10pt}
	\subfloat[Result w/ line correspondences.]{
	\includegraphics[width=0.45\textwidth]{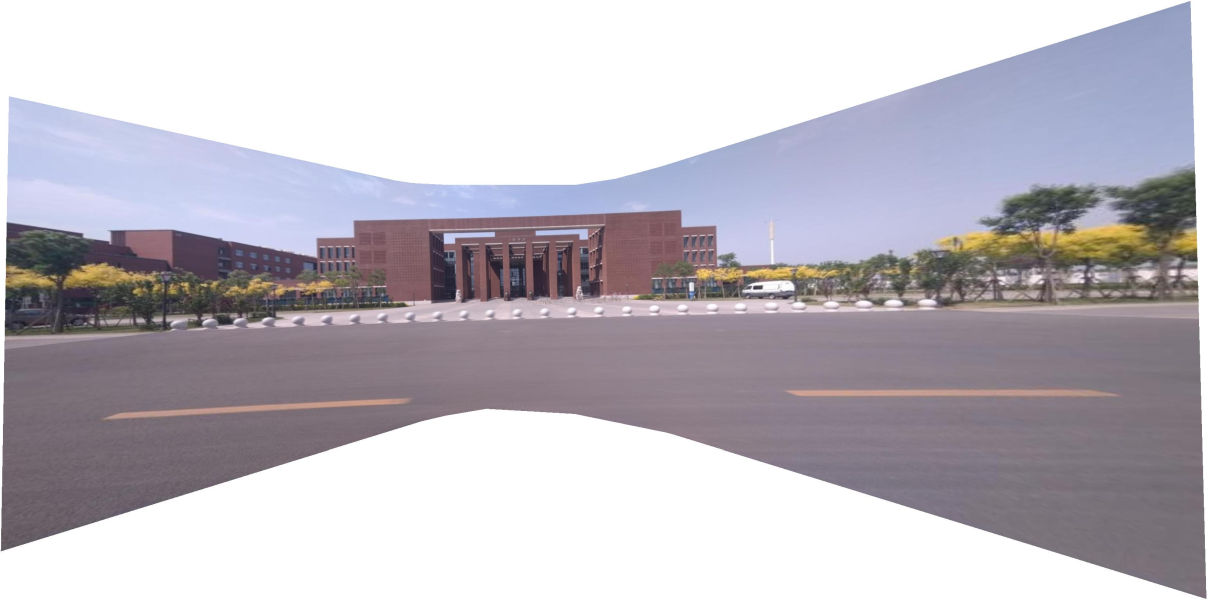}}
	\caption{Comparison of bundle adjustment results that without and with line correspondences for stitching ten images.}
	\label{fig:bundle}
\end{figure}

Given a sequence of multiple images, our stitching strategy
consists of three stages. In the first stage, we pick a reference
image as a standard perspective, such that other images should
be consistent with it. Then, we estimate a homography warp for
each image and transform them in the coordinate system of the
reference via bundle adjustment.
Our bundle adjustment method
is different from \cite{triggs1999bundle}, since we take the line correspondences as the supplement of the point correspondences, such that the result is more natural-looking (see Figure \ref{fig:bundle}). Finally,
we obtain the warped result for
each image via optimizing an energy function simultaneously.

\subsubsection{Bundle Adjustment}

For a set of input images $\{I_k\}_{k=1}^K$, we first pick a reference frame $I'$. For simplicity, $I'$ is chosen from one of $\{I_k\}_{k=1}^K$. Then we map all point and line features onto $I'$ via some homography chains. In the coordinate system of $I'$, only two endpoints of the line features are mapped and the coordinates with the same identity (both points and lines) are averaged. This process results in a set coordinates $\{\mathbf{\hat{p}}_i\}_{i=1}^N$ of point correspondences and a set coordinates $\{\mathbf{\hat{q}}^{s,e}_j\}_{j=1}^M$ of line correspondences in $I'$.
Finally, we simultaneously minimize the \textit{transfer error} of all point and line correspondences. Specifically, we  minimize the energy function
\begin{align}\label{eq_bundle_adjusment}
E(\Theta)&=\sum_{i=1}^N \frac{1}{\sum_{k=1}^K \delta_{ik}}\sum_{k=1}^K \delta_{ik}\|\mathbf{p}^k_i-f(\mathbf{x}_i, \mathbf{H}_k)\|^2\nonumber\\
 &+\sum_{j=1}^M \frac{1}{\sum_{k=1}^K \mu_{jk}}\sum_{k=1}^K \mu_{jk}\|\langle \vec{\mathbf{n}}^k_j,f(\mathbf{y}^{s,e}_j, \mathbf{H}_k)\rangle+c^k_j\|^2\nonumber\\
 &+\sum_{j=1}^M(\|\mathbf{y}^s_j-\mathbf{y}^e_j\|-c)^2,
\end{align}
where $\Theta=[\mathbf{H}_1,\dots,\mathbf{H}_K, \mathbf{x}_1,\dots,\mathbf{x}_N,\mathbf{y}^{s,e}_1,\dots,\mathbf{y}^{s,e}_M]$, $f(\mathbf{x}, \mathbf{H})$
is the homography that maps $\mathbf{x}\in I'$ into $I_k$ and $c$ is a positive constant that prevents $\mathbf{y}^{s,e}$ from degenerating into a single point.

In the initialization, we set $\mathbf{x}_i=\mathbf{\hat{p}}_i$ and $\mathbf{y}^{s,e}_j=\mathbf{\hat{q}}^{s,e}_j$.
The $k$-th homography $\mathbf{H}_k$ for image $I_k$ is initialized using DLT on the correspondences of points and lines between $I'$ and $I_k$. $\delta_{ik}$ (or $\mu_{jk}$) is an indicator that equals to one if the correspondence $\{\mathbf{p}^k_i, \mathbf{\hat{p}}_i\}$ (or $\{\mathbf{l}^k_j, \mathbf{\hat{q}}^{s,e}_j\}$) exists and otherwise it equals to zero. The division of each term in (\ref{eq_bundle_adjusment}) by $\sum_{k=1}^K \delta_{ik}$ (or $\sum_{k=1}^K \mu_{jk}$) guarantees that the points $\mathbf{\hat{p}}_i$ (or $\mathbf{\hat{q}}^{s,t}_j$) that are matched with many target images do not dominate. The Jacobian of (\ref{eq_bundle_adjusment}) is extremely sparse such that we can use the sparse Levenberg-Marquardt library of \cite{ceres-solver} to efficiently minimize it.

\subsubsection{Simultaneous Optimization}
After homographies $\{\mathbf{H}_k\}_{k=1}^K$
are settled via bundle adjustment, we can transform the perspective of each target image $I_k$ to that of the reference $I'$ by $\mathbf{H}_k^{-1}$, which is used to calculate the set of cross-line correspondences.

For multiple-image stitching, the alignment and naturalness terms are modified to
\begin{align}
E_{\mathrm{p}}(\hat{V},\hat{P})&=\sum_{i=1}^N \frac{1}{\sum_{k=1}^K \delta_{ik}}\sum_{k=1}^K \delta_{ik}\|\varphi(\hat{p}^k_i)-\hat{p}'_i\|^2\nonumber\\
					&=\|W_{\mathrm{p}}\hat{V}-\hat{P}\|^2\label{eq_multi-al},\\
E_{\mathrm{l}}(\hat{V},\hat{C}) & =\sum_{j=1}^M \frac{1}{\sum_{k=1}^K \mu_{jk}}\sum_{k=1}^K \mu_{jk}\|\langle \vec{\mathbf{n}}_j,\varphi(\hat{p}_{kj}^{s,e})\rangle+\hat{c}_j\|^2\nonumber\\
& =\|W_{\mathrm{l}}\hat{V}+\hat{C}\|^2\label{eq_multi-na},
\end{align}
where $\vec{\mathbf{n}}_j$ is the normal vector of the line with endpoints $\mathbf{\hat{y}}^{s,e}_j$ from bundle adjustment. $\hat{P},\hat{C}$ are intermediate variables that represent the point and line feature correspondences in $I'$.
For the distortion term, $\Omega$ is determined sequentially according to some homography chains. For the saliency term, the normal vectors of the corresponding lines are calculated from $\mathbf{H}_k^{-1}$.

Since the total energy function for multiple-image stitching is still sparse and quadratic, it can be efficiently minimized by any sparse linear solver.

\section{Experiments}\label{exp}

We demonstrate the effectiveness of our mesh-based warp in two aspects.
Firstly, we show a quantitative evaluation of the alignment accuracy for comparing our warp with two state-of-the-art single-perspective warps, homography and APAP~\cite{zaragoza2014projective}. Secondly, we
show a qualitative evaluation of the naturalness quality for comparing our warp with five state-of-the-art warps, homography, APAP, AutoStitch~\cite{Brown:2007}, SPHP~\cite{chang2014shape} and GSP~\cite{chen2016natural}. We also compare the time efficiency of our mesh-based warp with homography, APAP, SPHP, GSP and QH~\cite{li2017quasi}.

In the experiment, we use VLFeat~\cite{vedaldi2010vlfeat} to extract and match SIFT~\cite{lowe2004distinctive} features, use RANSAC~\cite{fischler1981random} to remove outliers and use LSD~\cite{von2010lsd} to detect line segments and match them by~\cite{jia2016novel}.
For the parameter setting, the grid size is set to $40\times 40$ for mesh deformation, $\lambda_{l}$, $\lambda_{ps}$, $\lambda_{pj}$ are set to 5, 50, 5 for energy minimization, which are relatively stable in the experiment. It is worth to note that the saliency term and the projective term are competitive. It means that if $\lambda_{\mathrm{s}}$ is too small then salient lines are obviously bent and if $\lambda_{\mathrm{s}}$ is too large then projective distortion become serious. In the experiment, we find $\lambda_{\mathrm{s}}=5$ is an appropriate trade-off (see Figure \ref{fig-visible}(b-d)).

Codes are implemented in MATLAB (some are in C++ for efficiency) and experiments are ran on a desktop PC with Intel i5 2.9GHz CPU and 8GB RAM.

\subsection{Quantitative Evaluation of Alignment}

We quantitatively evaluate the alignment accuracy of our proposed mesh-based warp, which is measured by the root mean squared error (RMSE) on a set of point correspondences $\{p_i,p'_i\}_{i=1}^N$,
\begin{equation}
 \mathrm{RMSE}(t)=\sqrt{\frac{1}{N}\sum_{i=1}^N\|t(p_i)-p'_i\|^2},
\end{equation}
where $t: \mathbb{R}^2\mapsto\mathbb{R}^2$ is a planar warp.

Similar to APAP, we randomly partition the available SIFT feature correspondences into ``training'' and ``testing'' sets that are of equal size. The training set is used to learn a warp and the RMSE is evaluated over both sets. We compare our warp with homography (Homo) and APAP (use the implementation provided by the authors).

\begin{table}
	\scriptsize
	\centering
	\caption{Average RMSE (TR: training set error, TE: testing set error)}
	\renewcommand\arraystretch{1.2}
	\begin{tabular}{|cc|c|c|c|}
		\hline
		\multicolumn{2}{|c|}{\multirow{2}[2]{*}{Dataset}} & Homo & APAP & Ours \\
		\multicolumn{2}{|c|}{~} & $\slash$Homo+DF & $\slash$APAP+DF & $\slash$Ours+DF \\
		\hline
		\textit{APAP-} & -TR   & 5.58  & 5.43  & \textbf{3.23 } \\
		\textit{railtracks} & -TE   & 5.69  & 5.51  & \textbf{3.76 } \\
		& -\% outliers & 15.44$\slash$15.33 & \textbf{13.09$\slash$13.12} & 14.03$\slash$13.76 \\
		\hline
		\textit{APAP-} & -TR   & 2.26  & 1.73  & \textbf{0.85 } \\
		\textit{garden} & -TE   & 2.29  & 1.75  & \textbf{1.06 } \\
		& -\% outliers & 14.59$\slash$15.00 & 14.31$\slash$14.32 & \textbf{14.27$\slash$14.29} \\
		\hline
		\textit{APAP-} & -TR   & 5.97  & 5.09  & \textbf{4.40 } \\
		\textit{conssite} & -TE   & 6.30  & 5.83  & \textbf{5.46 } \\
		& -\% outliers & 9.12$\slash$8.43 & 7.68$\slash$7.92 & \textbf{7.53$\slash$7.07} \\
		\hline
		\textit{SPHP-} & -TR   & 1.81  & 1.79  & \textbf{1.35 } \\
		\textit{garden} & -TE   & 1.82  & 1.81  & \textbf{1.54 } \\
		& -\% outliers & 21.75$\slash$21.61 & 21.07$\slash$21.07 & \textbf{20.79$\slash$20.99} \\
		\hline
		\textit{SPHP-} & -TR   & 2.07  & 1.94  & \textbf{1.57 } \\
		\textit{street} & -TE   & 2.15  & 2.08  & \textbf{1.89 } \\
		& -\% outliers & 29.66$\slash$29.74 & 29.43$\slash$29.47 & \textbf{29.37$\slash$29.31} \\
		\hline
		\textit{DH-} & -TR   & 7.84  & 5.95  & \textbf{4.84 } \\
		\textit{temple} & -TE   & 7.81  & 6.11  & \textbf{5.40 } \\
		& -\% outliers & 13.27$\slash$13.33 & \textbf{11.27}$\slash$10.72 & 12.07$\slash$\textbf{10.52} \\
		\hline
		\textit{DH-} & -TR   & 4.90  & 3.93  & \textbf{1.37 } \\
		\textit{carpark} & -TE   & 4.95  & 4.06  & \textbf{2.05 } \\
		& -\% outliers & 16.66$\slash$16.46 & 13.20$\slash$13.07 & \textbf{13.19$\slash$12.73} \\
		\hline
		\textit{SVA-} & -TR   & 4.08  & 3.25  & \textbf{2.92 } \\
		\textit{chess/girl} & -TE   & 4.08  & 3.30  & \textbf{3.20 } \\
		& -\% outliers & 21.34$\slash$21.40 & \textbf{20.66$\slash$20.60} & 21.01$\slash$20.67 \\
		\hline
		\textit{AANAP-} & -TR   & 9.63  & 9.43  & \textbf{4.16 } \\
		\textit{building} & -TE   & 10.92  & 10.89  & \textbf{6.48 } \\
		& -\% outliers & 57.56$\slash$57.87 & 57.14$\slash$57.12 & \textbf{57.05$\slash$57.01} \\
		\hline
	\end{tabular}%
	\label{tab-error}%
\end{table}%

We also employed the pixel-wise error metric in~\cite{lin2012aligning}, where a pixel $x$ in $I$ is labeled as an outlier if there is no similar pixel (intensity difference less than ten gray levels) within the four-pixel neighborhood of $t(x)$ in $I'$. Then the percentage of outliers in the overlapping region resulting from $t$ is regarded as the warping error. We evaluate the \% outliers for three warps without and with dual-feature, which are showed before and after ``$\slash$''.

Table~\ref{tab-error} shows the average RMSE and \% outliers over twenty repetitions on nine challenging image pairs, which come from APAP, SPHP, DH, SVA and AANAP. It is clear that our warp yields the lowest errors for most pairs (bold values).

\subsection{Qualitative Evaluation of Naturalness}

\begin{figure*}
	\centering
	\subfloat[AutoStitch]{
		\includegraphics[width=0.99\textwidth]{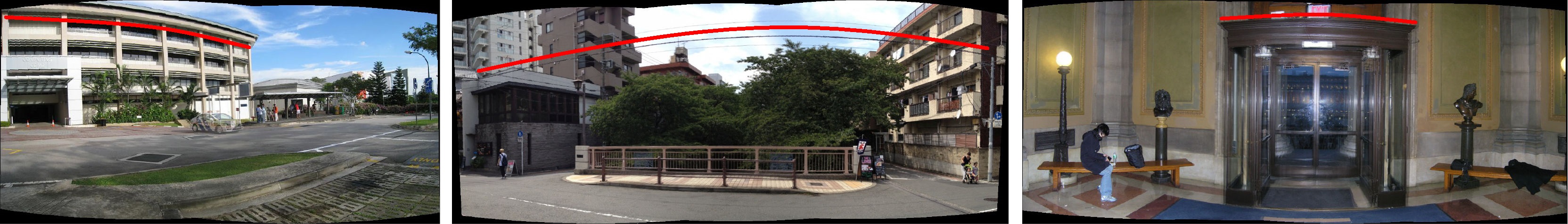}}\\
	\vspace{-5pt}
	\subfloat[SPHP]{
		\includegraphics[width=0.99\textwidth]{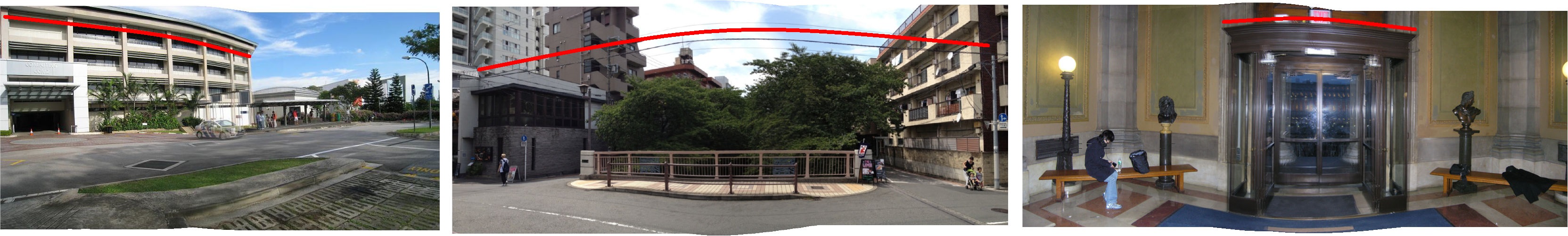}}\\
	\vspace{-5pt}
	\subfloat[GSP]{	
		\includegraphics[width=0.99\textwidth]{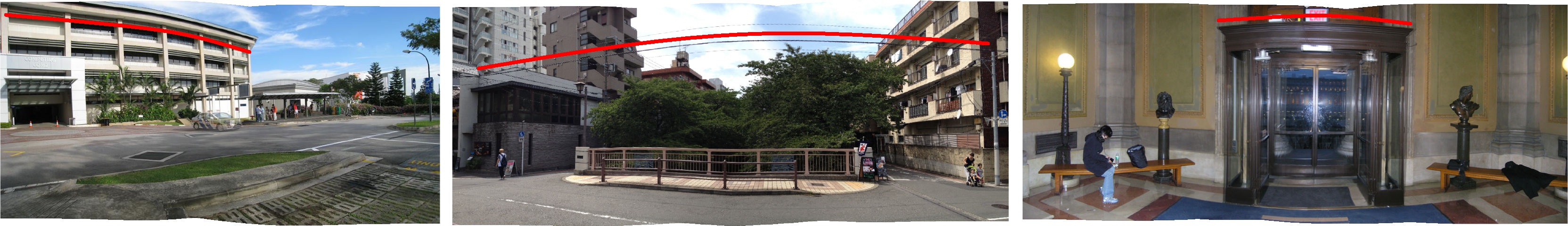}}\\
	\vspace{-5pt}
	\subfloat[Homo]{
		\includegraphics[width=0.99\textwidth]{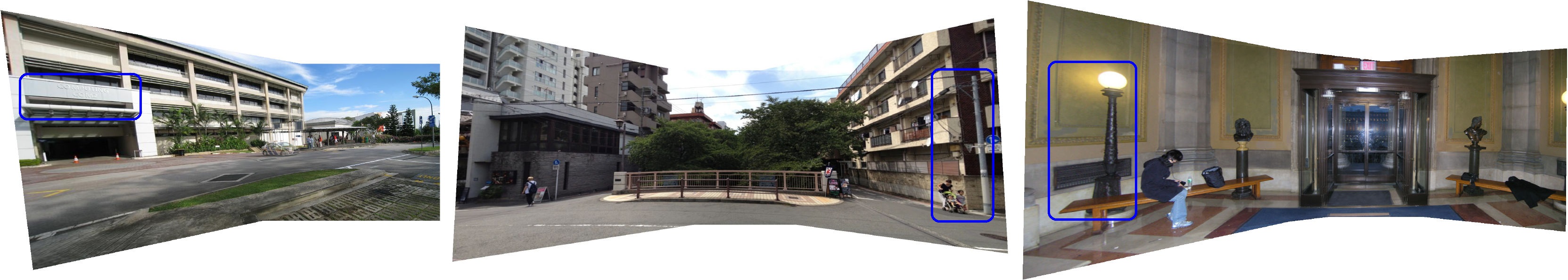}}\\
	\vspace{-5pt}
	\subfloat[APAP]{
		\includegraphics[width=0.99\textwidth]{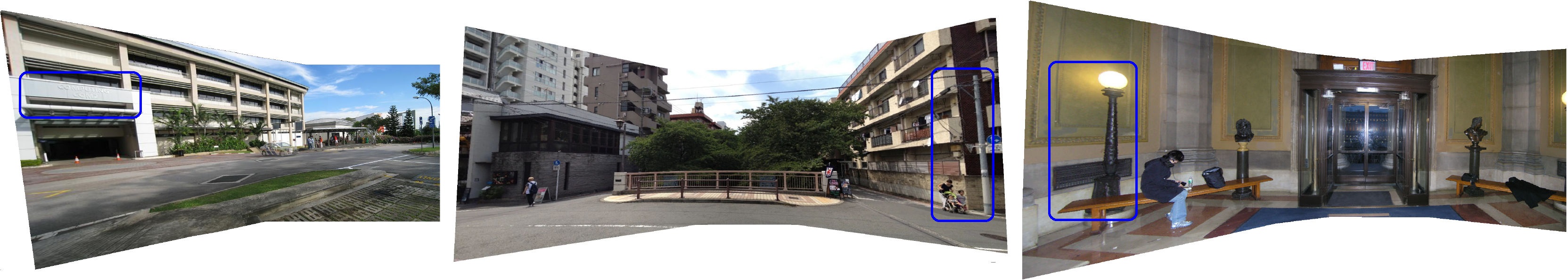}}\\
	\vspace{-5pt}
	\subfloat[Ours]{
		\includegraphics[width=0.99\textwidth]{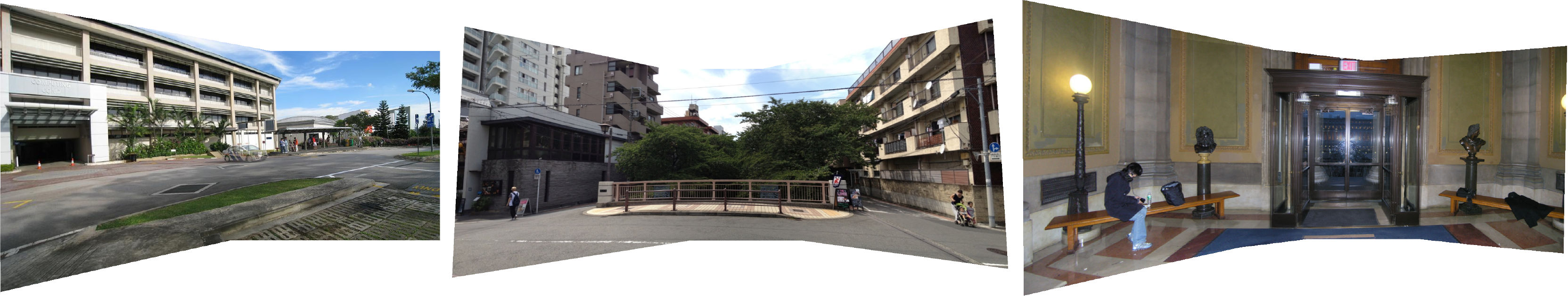}}\\
	\caption{Three comparison results for stitching 2, 3 and 4 images. From \textbf{Left} to \textbf{Right}, input images are from \cite{chang2014shape}, \cite{gao2011constructing} and \cite{nomura2007scene} respectively. (Best to zoom-in and view on screen)}
	\label{fig:comp1}
\end{figure*}

\begin{figure*}
	\centering
	\subfloat[AutoStitch]{
		\includegraphics[width=0.78\textwidth]{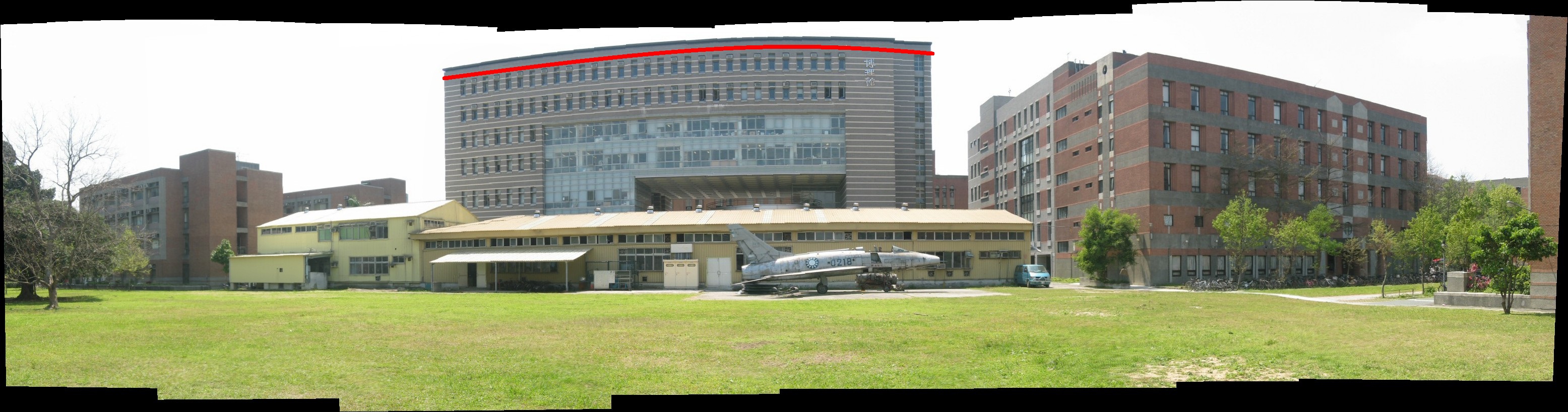}}\\
	\vspace{-5pt}
	\subfloat[GSP]{	
		\includegraphics[width=0.78\textwidth]{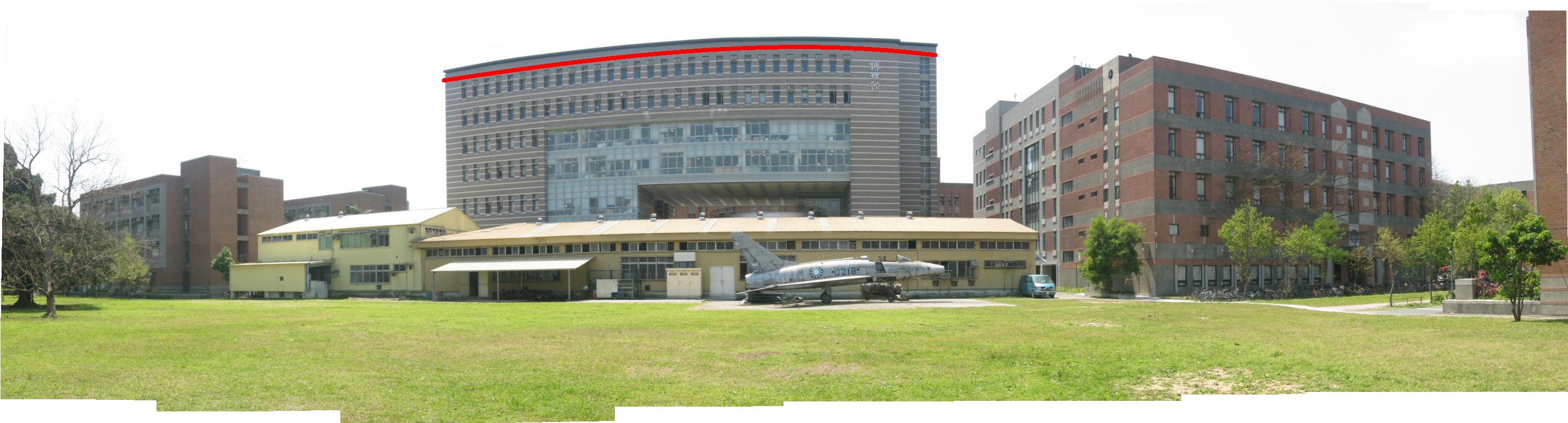}}\\
	\vspace{-5pt}
	\subfloat[Homo]{
		\includegraphics[width=0.78\textwidth]{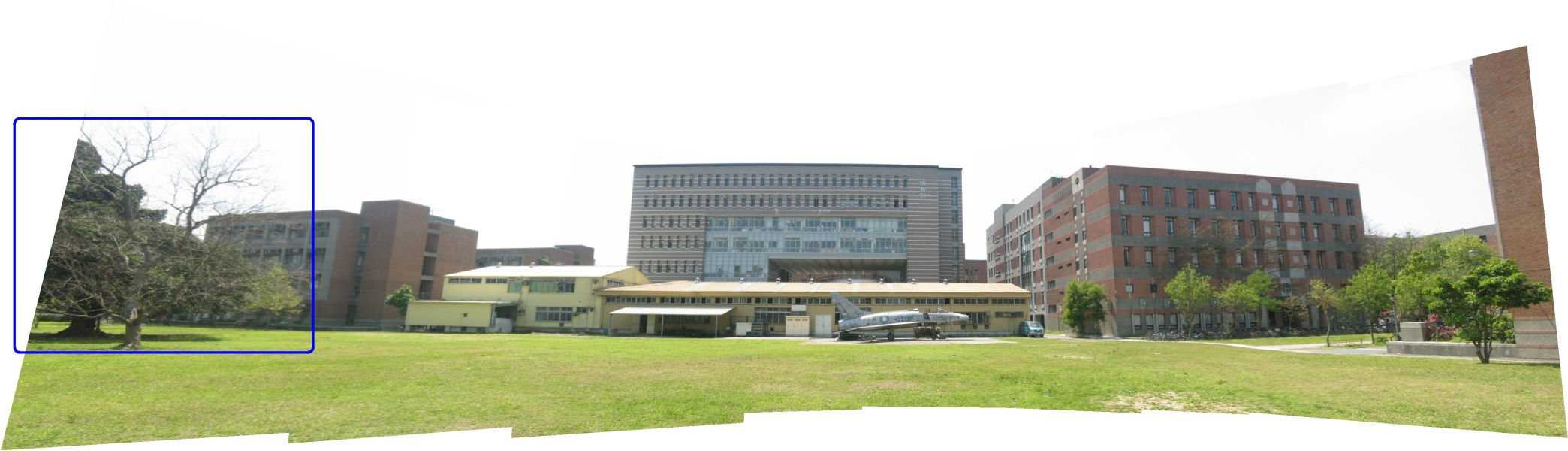}}\\
	\vspace{-5pt}
	\subfloat[APAP]{
		\includegraphics[width=0.78\textwidth]{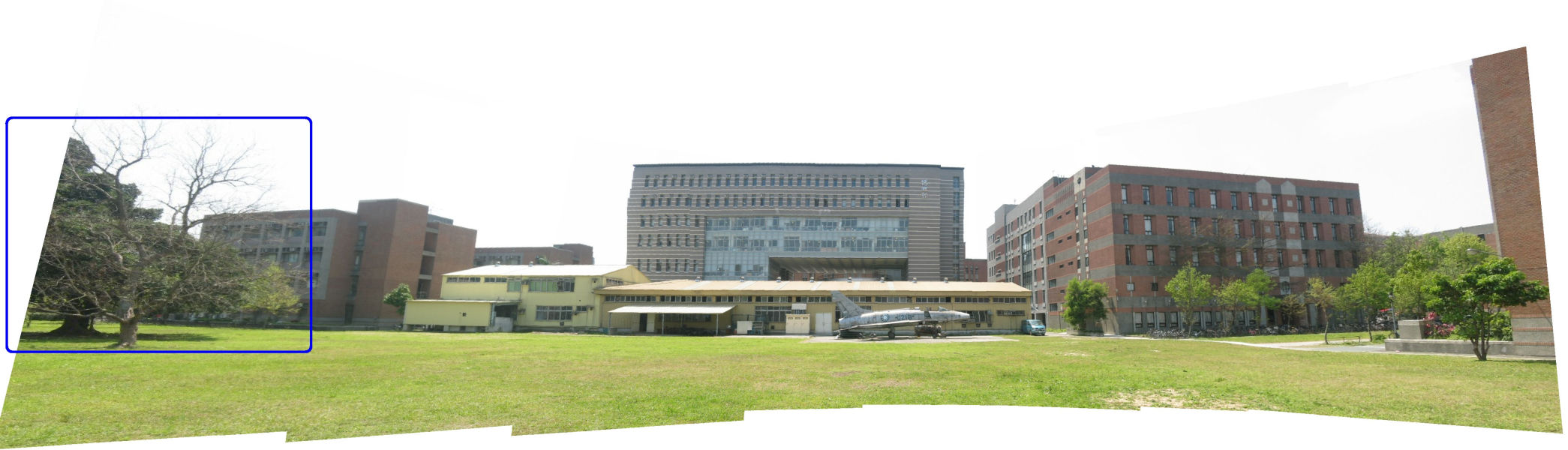}}\\
	\vspace{-5pt}
	\subfloat[Ours]{
		\includegraphics[width=0.78\textwidth]{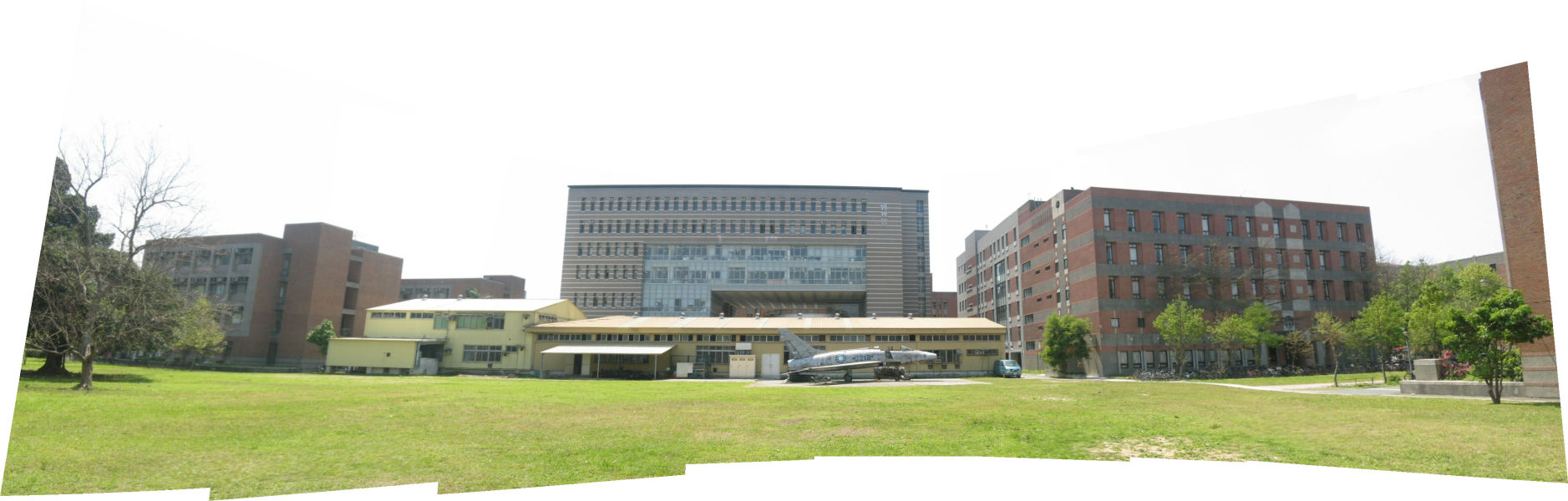}}\\
	\caption{A comparison result for stitching $11$ images. SPHP fails to output any result due to the wide field of view. Input images are from \cite{chen2016natural}.}
	\label{fig:comp2}
\end{figure*}

We qualitatively compare the naturalness quality of our proposed mesh-based warp with homography, APAP, AutoStitch, SPHP and GSP. Figure \ref{fig:comp1} and \ref{fig:comp2} demonstrate four comparison results where input images are from some open datasets.
It is clear that our warp creates more natural-looking stitching results in aspects of naturalness, distortion and saliency. More comparison results of overall performance from our dataset and other datasets including DH, CAVE \cite{nomura2007scene}, SPHP, AANAP, GSP and APAP are available in the supplementary material.

\begin{figure}
	\centering
	\subfloat[Residual-iteration curve.]{
		\includegraphics[height=0.135\textheight]{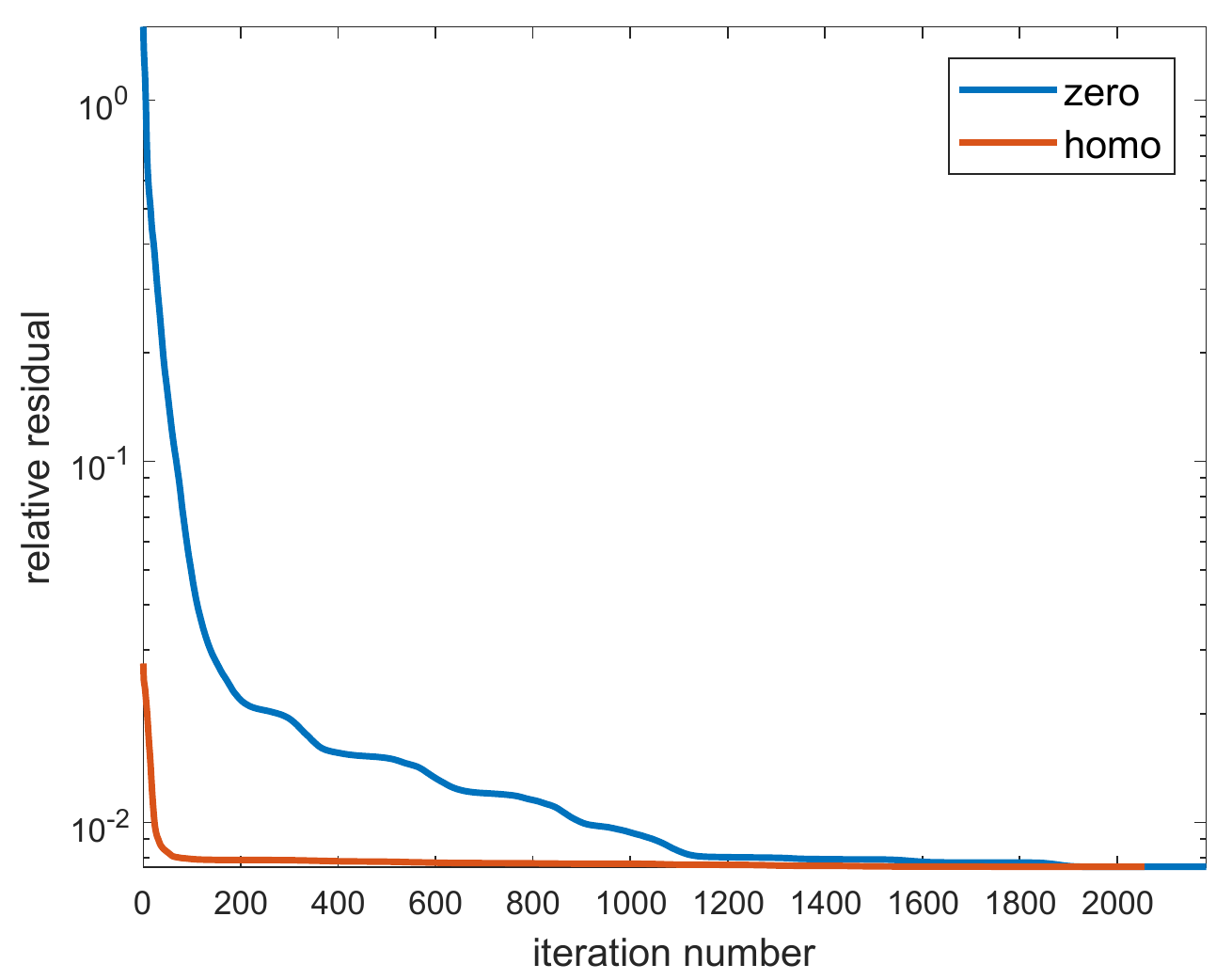}}
	\subfloat[Elapsed time.]{
		\includegraphics[height=0.14\textheight]{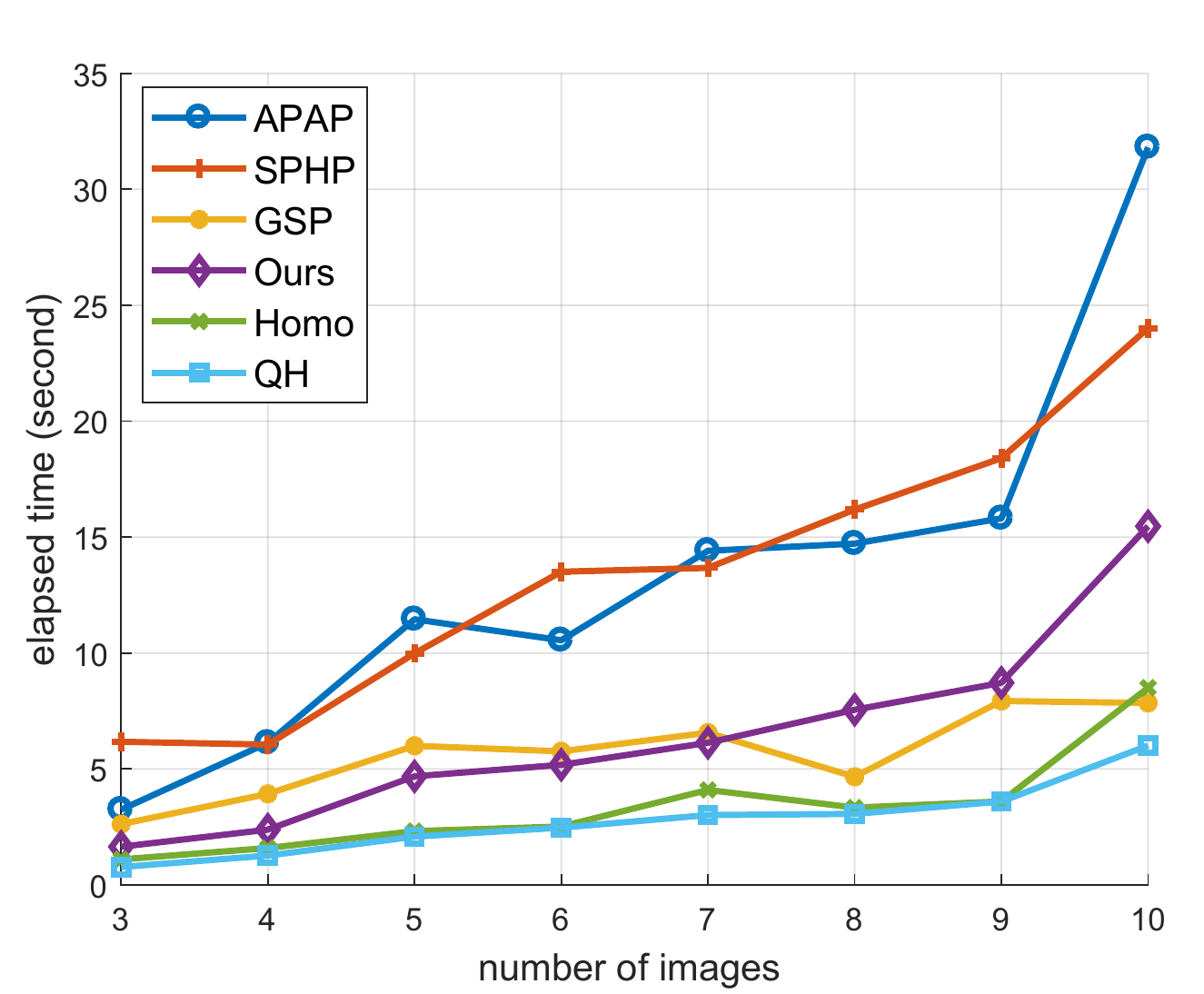}}
	\caption{(a) Residual-iteration curve of our method for different initializations. (b) Elapsed time of different methods for stitching multiple images.}
	\label{fig:time}
\end{figure}

\subsection{Time Efficiency}

We compare the time efficiency of our proposed mesh-based warp with some publicly available warps, homography,  APAP, SPHP, GSP and QH. All the mesh-based methods are ran with the same grid size.

Figure \ref{fig:time}(a) shows the residual-iteration curve of our warp for different initializations.
The convergence is much faster for the initialization of homography.

Figure \ref{fig:time}(b) illustrates the elapsed time of different methods for stitching multiple images. Except GSP is running in C++, all the other methods are running in MATLAB. The recorded time includes mesh optimization, texture mapping and linear blending, where feature detecting and matching are not taken into account. It is obvious that homography and QH are the most efficient warps. APAP is the least efficient warp because it calculates many local homography warps. GSP and our warp are comparative because they both solve the mesh deformation simultaneously.

\subsection{Failure Cases}
Our proposed warp could fail if the parallax is too large in the overlapping region, such that
the common dominant plane cannot well represent the perspective transdormation between the images.

\section{Conclusion}

In this paper, we proposed two single-perspective warps for image stitching such that the stitching results look as natural as possible.
The first one was a parametric warp, which is a combination of APAP and QH via DF.
The second one was a mesh-based warp,
which is determined by
optimizing a sparse and quadratic total energy function.
A comprehensive evaluation demonstrated that
	the proposed warp outperforms some state-of-the-art
	warps, including homography, APAP, AutoStitch, SPHP and GSP.

\ifCLASSOPTIONcaptionsoff
  \newpage
\fi




\end{document}